\documentclass[letterpaper, 10 pt, journal, twoside]{IEEEtran}

\ifCLASSINFOpdf

\else

\fi

\usepackage{graphics} %
\usepackage{epsfig} %

\usepackage{cite} %
\usepackage{amsmath}
\usepackage{amssymb} %
\usepackage{subcaption}
\usepackage{graphicx}
\usepackage{todonotes}
\usepackage{makecell}
\usepackage[pdfa]{hyperref}
\usepackage{multirow}
\usepackage{ctable}

\usepackage[bottom]{footmisc} %

\usepackage[normalem]{ulem}
\useunder{\uline}{\ul}{}

\usepackage{float}
\usepackage{dblfloatfix} %

\usepackage{xsavebox} 
\usepackage{atbegshi} 

\xsavebox{PageBGPicture}{
	\begin{tikzpicture}
	\node [rectangle, minimum width=15cm, minimum height=1.5cm, align=center, text = gray] (box){
	 \fontsize{7.7}{8}\selectfont
	 \copyright 2021 IEEE. Personal use of this material is permitted.  Permission from IEEE must be obtained for all other uses, in any current or future media, including reprinting/republishing this material \\
	 \fontsize{7.7}{8}\selectfont 
	 for advertising or promotional purposes, creating new collective works, for resale or redistribution to servers or lists, or reuse of any copyrighted component of this work in other works.\\
	};
	
	\end{tikzpicture}
	
}

\AtBeginShipout{
	\AtBeginShipoutUpperLeft{\raisebox{-\height}{\xusebox{PageBGPicture}}}
}

\begin{document}

\title{\LARGE \bf The GRIFFIN Perception Dataset: Bridging the Gap Between Flapping-Wing Flight and Robotic Perception}

\author{J.P. Rodr\'{\i}guez-G\'omez, R. Tapia, J. L. Paneque, P. Grau, A. G\'omez Egu\'{\i}luz,\\ J.R. Mart\'{\i}nez-de Dios and A. Ollero
\thanks{Manuscript received October 15, 2020; Revised November 30, 2020; Accepted January, 21, 2021.}
\thanks{This paper was recommended for publication by Cesar Cadena upon evaluation of the Associate Editor and Reviewers’ comments.}
\thanks{The authors are with the GRVC Robotics Lab Sevilla. Universidad de Sevilla, Spain %
{\tt email: \{jrodriguezg, raultapia, jlpaneque, pgrau, ageguiluz, jdedios, aollero\}@us.es}. This work was supported by the European Research Council as part of GRIFFIN ERC Advanced Grant (Action 788247) (\url{https://griffin-erc-advanced-grant.eu}), the  European  Commission as part of AERIAL-CORE project (Grant H2020-2019-871479), and ARM-EXTEND (DPI2017-8979-R) funded by the Spanish National R\&D Plan.}
\thanks{Digital Object Identifier (DOI): see top of this page.}
}

\markboth{IEEE Robotics and Automation Letters. Preprint Version. Accepted JANUARY, 2021}
{Rodriguez-Gomez \MakeLowercase{\textit{et al.}}: The GRIFFIN Perception Dataset}

\maketitle

\begin{abstract}
The development of automatic perception systems and techniques for bio-inspired flapping-wing robots is severely hampered by the high technical complexity of these platforms and the installation of onboard sensors and electronics. Besides, flapping-wing robot perception suffers from high vibration levels and abrupt movements during flight, which cause motion blur and strong changes in lighting conditions. This paper presents a perception dataset for bird-scale flapping-wing robots as a tool to help alleviate the aforementioned problems. The presented data include measurements from onboard sensors widely used in aerial robotics and suitable to deal with the perception challenges of flapping-wing robots, such as an event camera, a conventional camera, and two Inertial Measurement Units (IMUs), as well as ground truth measurements from a laser tracker or a motion capture system. A total of 21 datasets of different types of flights were collected in three different scenarios (one indoor and two outdoor). To the best of the authors' knowledge this is the first dataset for flapping-wing robot perception.
\end{abstract}


\IEEEpeerreviewmaketitle

\section*{Supplementary Material}

The dataset is available at: \url{http:\\grvc.us.es/eye-bird-dataset}

\section{Introduction}
\label{sec:intro}

In the last years, there has been a growing interest in the development of bio-inspired aerial robots. The potential advantages of these platforms, such as flapping-wing robots, over traditional rotary-wing and fixed-wing platforms in terms of energy consumption and safety in populated environments have motivated significant R\&D efforts that have resulted in the development of bird-scale platforms (i.e. ornithopters), such as \cite{folkertsma2017robird}, \cite{festo2011smartbird}, or \cite{ryu2020design}, as well as insect-scale platforms \cite{farrell2018review}.
However, very few of the reported flapping-wing platforms include onboard perception capabilities. We are interested in perception for bird-scale flapping-wing robots, also called ornithopters, which have enough payload capability for onboard sensors and embedded computers. 
This work  
has been developed within the GRIFFIN ERC Advanced Grant, which is devoted to the development of ornithopter robots capable of navigating, maneuvering, perching, and manipulating objects in an autonomous manner.

Flapping-wing flight suffers from mechanical vibrations, and wide abrupt movements caused by the lift and thrust generated during downward strokes, which cause limitations in existing perception techniques \cite{gomezeguiluz2019towards}. Research in flapping-wing robot perception is constrained by the high complexity of these platforms. Besides, although there are a number of datasets for aerial robot perception, none of them provide sensor data collected during flapping-wing flights.

\begin{figure}[t]
    \begin{center}
    \includegraphics[width=0.9\linewidth]{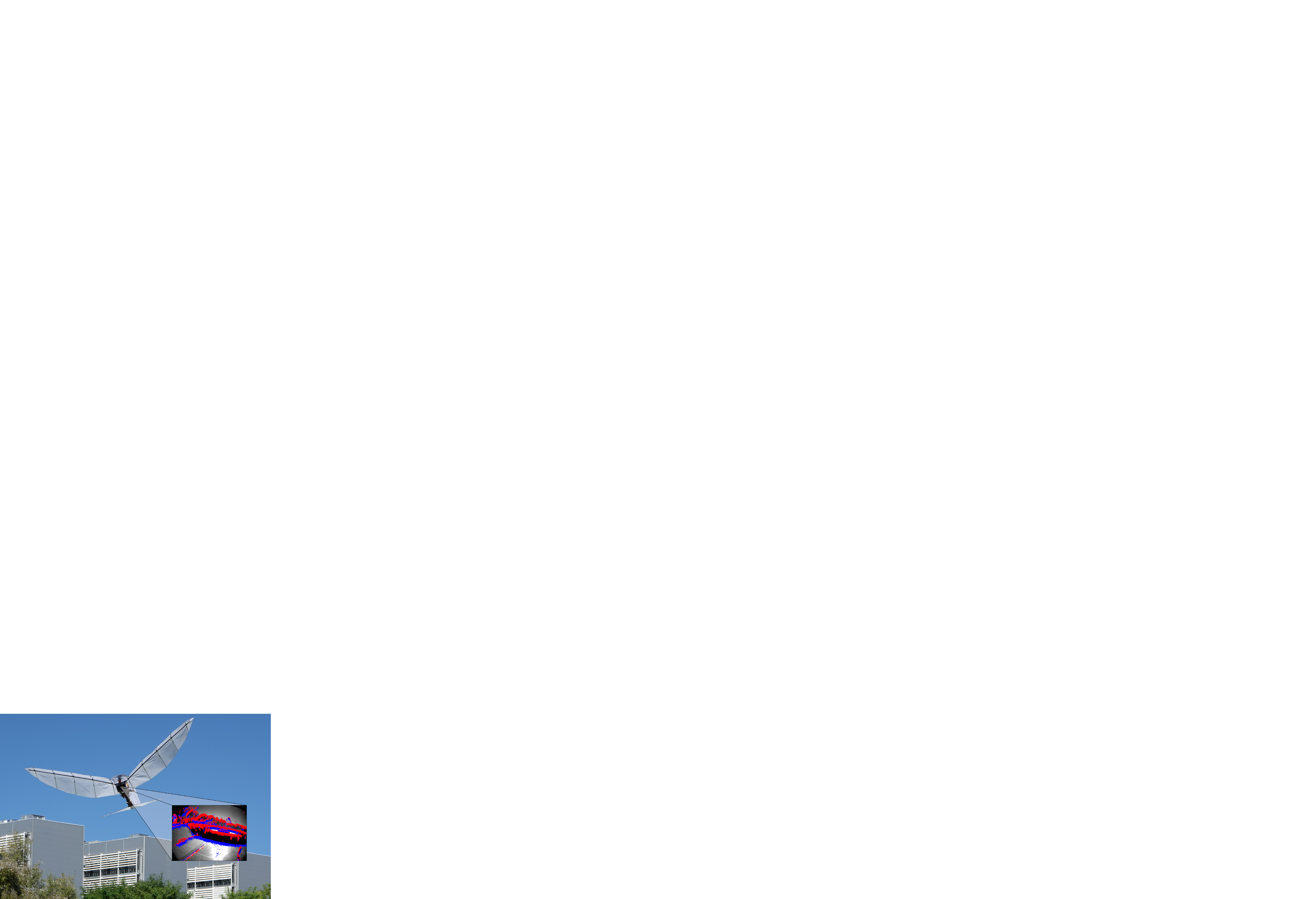}
    \caption{The GRIFFIN \textit{Eye-Bird} ornithopter robot during a dataset recording flight.} 
    \label{fig:powebird}
    \end{center}
    \vspace{-5mm}       
\end{figure}

\begin{table*}[!]
\centering
\scalebox{1.0}{
\begin{tabular}{c | c c c c c }
\specialrule{.1em}{.05em}{.05em} 
\textbf{Dataset} & \textbf{Platform} & \textbf{Event  camera} & \textbf{Data type} & \textbf{Scenario} \\ \hline \hline
Mid Air \cite{fonder2019mid}                                                             & Multirotor                                                                                                                                 & No                                                                                                       & Synthetic                                                          & Outdoors                                                                    \\ 
AU-AIR \cite{bozcan2020auair}                                                              & Multirotor                                                                                                                               & No                                                                                                   & Real                                  & Outdoors                                                             \\ 
V4RL Wide-baseline Place Recognition \cite{maffra2019real}                                                             & Multirotor                                                                                                                                 & No                                                                                                       &  Synthetic \& Real                                                       & Outdoors                                                                    \\ 
EuRoC \cite{burri2016euroc}                                                               & MAV                                                                                                                                      & No                                                                                                    & Real                                                               & Indoors                                                              \\ 
Zurich Urban\cite{majdik2017zurich}                                                              & MAV                                                                                                                               & No                                                                                                  & Real                                                               & Outdoors                                                                    \\ 
Blackbird \cite{antonini2020blackbird}                                                           & MAV                                                                                                                              & No                                                                                                     & Synthetic \& Real                                                               & Indoors                                                                    \\ 
UPenn Fast Flight \cite{sun2018robust}                                                                & \begin{tabular}[c]{@{}c@{}}Multirotor drone racing\end{tabular}                                        & No                                                                                                 & Real                                                                  & Outdoors                                                                    \\ 
\begin{tabular}[c]{@{}c@{}}Multivehicle Stereo Event Camera \cite{zhu2018multivehicle}  \end{tabular}                                                            & Multirotor    & Yes                                                             & Real                                                                  & 
 
\begin{tabular}[c]{@{}c@{}}Indoors \&  outdoors\end{tabular}                                                                  \\ 
\begin{tabular}[c]{@{}c@{}}UZH-FPV Drone Racing \cite{delmerico2019we}    \end{tabular} 
                                                           & Multirotor drone racing                                                                                                                                                                    & Yes                                                             & Real                                                                  & \begin{tabular}[c]{@{}c@{}}Indoors \&  outdoors\end{tabular}                                                                     \\ 
ROSS-LAN \cite{rodriguez2019roslan}                                                           & \begin{tabular}[c]{@{}c@{}}Bio-inspired bird trajectory\end{tabular}                                                                            & Yes                                                                                                                        & Synthetic                                                          & Indoors                                                                    \\ \hline \hline
\textbf{GRIFFIN Perception} & \textbf{\begin{tabular}[c]{@{}c@{}}Ornithopter gliding\\ and flapping\end{tabular}}                                                            & \textbf{Yes}                                                                                              & \textbf{Real}                                                      & \textbf{Indoors \& Outdoors}                                                    \\ 
\specialrule{.1em}{.05em}{.05em} 
\end{tabular}
}
\caption{Summary of main reported datasets for aerial robot perception.}
\label{table:soa}
\end{table*}

This paper presents a perception dataset for ornithopter robots. It contains measurements gathered from onboard lightweight sensors commonly used in aerial robots and suitable to address the main challenges of perception on board flapping-wing robots: an event camera, a conventional camera, and IMUs. Event cameras have high temporal resolution and dynamic range, which make them very robust against motion blur and lighting conditions. Traditional cameras and IMUs are two of the most widely used sensors in aerial robotics. Besides, the datasets include ground truth measurements from a laser tracker (in outdoor scenarios, see Figure \ref{fig:powebird}) or a motion capture system (in indoors), and ArUco visual markers scattered in the scenario. Three different types of datasets are included: 1) \emph{Base} datasets, with agile maneuvers that exploit the ornithopter flying capability; 2) \emph{ArUco} datasets, with ArUco markers in the scenario mapped with ground truth positioning; and 3) \emph{People} datasets, which provide samples for developing and evaluating object detection techniques based on events and/or visual images. The provided datasets can be useful to develop, tune or evaluate a wide range of image and event-based perception techniques from feature extraction and odometry to object detection, semantic segmentation or SLAM, among others.

The rest of the paper is organized as follows. Section \ref{sec:rw} summarizes the main works in the topics addressed in the paper. The flapping-wing platform and sensors used to collect the dataset are presented in Section \ref{sec:platform}. The sensor data format is given in Section \ref{subsec:dataset}.
Section \ref{sec:experimental} describes the dataset collection and evaluation. Finally, Section  \ref{sec:conclusion} concludes the paper and highlights the main future research steps.

\section{Related Work}
\label{sec:rw}

In recent years, a wide variety of drone datasets have been introduced for robotic perception tasks, see e.g., \cite{fonder2019mid, maffra2019real, bozcan2020auair, burri2016euroc, majdik2017zurich, antonini2020blackbird, sun2018robust}. 
The work in \cite{fonder2019mid} 
presents \textit{Mid-Air}, a large synthetic dataset of quadcopters flying in unstructured environments that 
includes data from different scenarios and climate conditions using the Airsim simulator.
The work in \cite{maffra2019real}
provides
a dataset 
including
real and photo-realistic synthetic data to evaluate place recognition methods with respect to viewpoint tolerance.
The \textit{AU-AIR} dataset \cite{bozcan2020auair} 
presents
annotated data of traffic surveillance onboard an Unmanned Aerial Vehicle (UAV) in addition to data from camera, GPS, and IMU. A summary of the main reported datasets for aerial robot perception is shown in Table \ref{table:soa}.

Some of the released datasets focus specifically on Micro Aerial Vehicles (MAVs) and multirotor drone racing motivated by the challenging perception issues posed by aggressive flights. 
The \textit{EuRoC} dataset  \cite{burri2016euroc} 
provides
data of MAV flights in two different indoor scenarios: an industrial scenario for evaluating visual-inertial localization, and a room equipped with a motion capture system for evaluating 3D reconstruction techniques. The MAV dataset 
in 
\cite{majdik2017zurich} 
includes flights within the urban streets of Zurich, Switzerland. The \textit{Blackbird UAV dataset} \cite{antonini2020blackbird} 
approaches the problem of agile and autonomous operation of aerial vehicles in outdoor environments with special emphasis on visual inertial navigation, 3D reconstruction, and depth estimation. The data from real flights 
has been extended by generating additional synthetic data through simulation. A dataset of fast flights using a quadrotor in an outdoor scenario 
is presented in \cite{sun2018robust}. Four different flights were
performed in an airport runway repeating the same trajectory at four different speeds 
in the range [$5$,$17.5$] m/s.
All the above datasets 
include
high-resolution images, GPS, and IMU data. Some of them also 
provide
additional measurements, such as stereo vision data (the  \textit{EuRoC} dataset) or rotor tachometers data, and depth images (the \textit{Blackbird UAV} dataset). Although the above datasets provide measurements relevant for perception techniques for multirotors and MAVs, they are not adequate for testing and developing techniques that deal with the perception difficulties raised during flapping.

Flapping-wing robots entail specific perception challenges and require perception systems and techniques that consider the effects of generating lift and thrust through wing flapping. One of the first approaches to cope with the challenges of ornithopter perception 
has been presented in \cite{pan2020development}. The authors
propose a vision-based stabilizing system to address the pitch and roll fluctuations during each flapping period. The system could be carried within the $<$100 g payload limitation of their ornithopter. Recently, some datasets useful for flapping-wing robot development have been published. In a previous work we
 presented
\textit{ROSS-LAN} \cite{rodriguez2019roslan}, a simulation scheme to obtain synthetic sensor data from trajectories that mimic bird flying maneuvers, and 
released 
some synthetic datasets. Also, 
the work in \cite{maldonado2020adaptive} 
includes experimental control data of an ornithopter performing landing manoeuvres in a scenario equipped with a motion capture system, but does not provide onboard sensing data that could be used in perception. To the best of our knowledge, no dataset with experimental onboard measurements  suitable for flapping-wing robot perception has been reported.

The strict payload capacity and weight distribution constraints together with the vibration level and abrupt movements of flapping-wing platforms require a careful selection of the sensors mounted onboard an ornithopter. In work \cite{gomezeguiluz2019towards} the suitability of LIDARs, conventional, and event cameras for ornithopters was analyzed, concluding that event-based vision provides a promising solution to many of these perception challenges.
Event cameras capture the illumination changes 
in the form of events with microsecond time resolution and high dynamic range.
Unlike conventional cameras, event cameras are robust to motion blur and lighting conditions. They have moderated weight, size, and low power consumption. The use of event-based vision onboard UAVs has received increasing research interest \cite{gallego2019event} in problems such as visual servoing \cite{gomezeguiluz2020async}, motion segmentation \cite{mitrokhin2019ev}, surveillance tasks
\cite{rodriguez2020async}, robot localization \cite{vidal2018ultimate}, and onboard computation load management \cite{tapia2020asap}, among many others.

A number of datasets for event-based vision have been presented exploring the advantages of event cameras onboard aerial robots \cite{rodriguez2020async,mitrokhin2018event, delmerico2019we, zhu2018multivehicle}.
The works in \cite{rodriguez2020async} and \cite{mitrokhin2018event}
provide sequences recorded onboard multirotors used to evaluate event-based methods for tracking moving objects. The \textit{Multivehicle Stereo Event Camera Dataset}, presented in \cite{zhu2018multivehicle}, 
also includes measurements from LIDAR and several IMUs. Among the vehicles used, an hexacopter 
provides sensor measurements under different perspectives, vehicle velocities, and illumination conditions. 
The work in \cite{delmerico2019we}
presents a dataset for autonomous drone racing 
including measurements from event cameras, conventional stereo-pair cameras, IMU, and the ground truth pose. However, none of these works have explored the use of event-based vision onboard flapping-wing robots.

\section{The Experimental Platform} 
\label{sec:platform}

This section presents the bird-scale flapping-wing robotic platform, the onboard sensors and the ground truth instruments used in the presented datasets.

\subsection{The GRIFFIN Eye-Bird Ornithopter}
\label{subsec:e-bird}

The ornithopter used 
in this work is an evolution of \emph{E-Flap}, a custom design developed by the GRVC Robotics Lab in the GRIFFIN project, which was modified with onboard sensors and electronics for perception research.

Flapping-wing aircrafts have strict restrictions in payload capacity and weight distribution. Adding payload to an ornithopter would make it demand more lift and thrust during flight, requiring higher flapping frequencies that induce higher stresses over the structural and mechanical parts. With a 1.5 m wingspan and an empty weight of 450 g, the design of \textit{Eye-Bird} was optimized for maximum payload capacity and ease of payload integration, with multiple attachment points though simple nuts, bolts, or cable ties over the body. A total of 250 g was added as payload (sensors, electronics, batteries, and cables) for the presented 
flights. Being more than half the empty weight of the ornithopter, its location strongly affects the center of gravity (CoG) and inertias, and hence, stability and maneuverability. The ornithopter total length is 95 cm. The 39 cm length tail brings the neutral point of the aircraft back to 25 cm from the head, which dictates the limit at which the CoG can be pulled back to keep the ornithopter stable.

\begin{figure}[t]
    \begin{center}
    \includegraphics[width=0.88\linewidth, height=6.3cm]{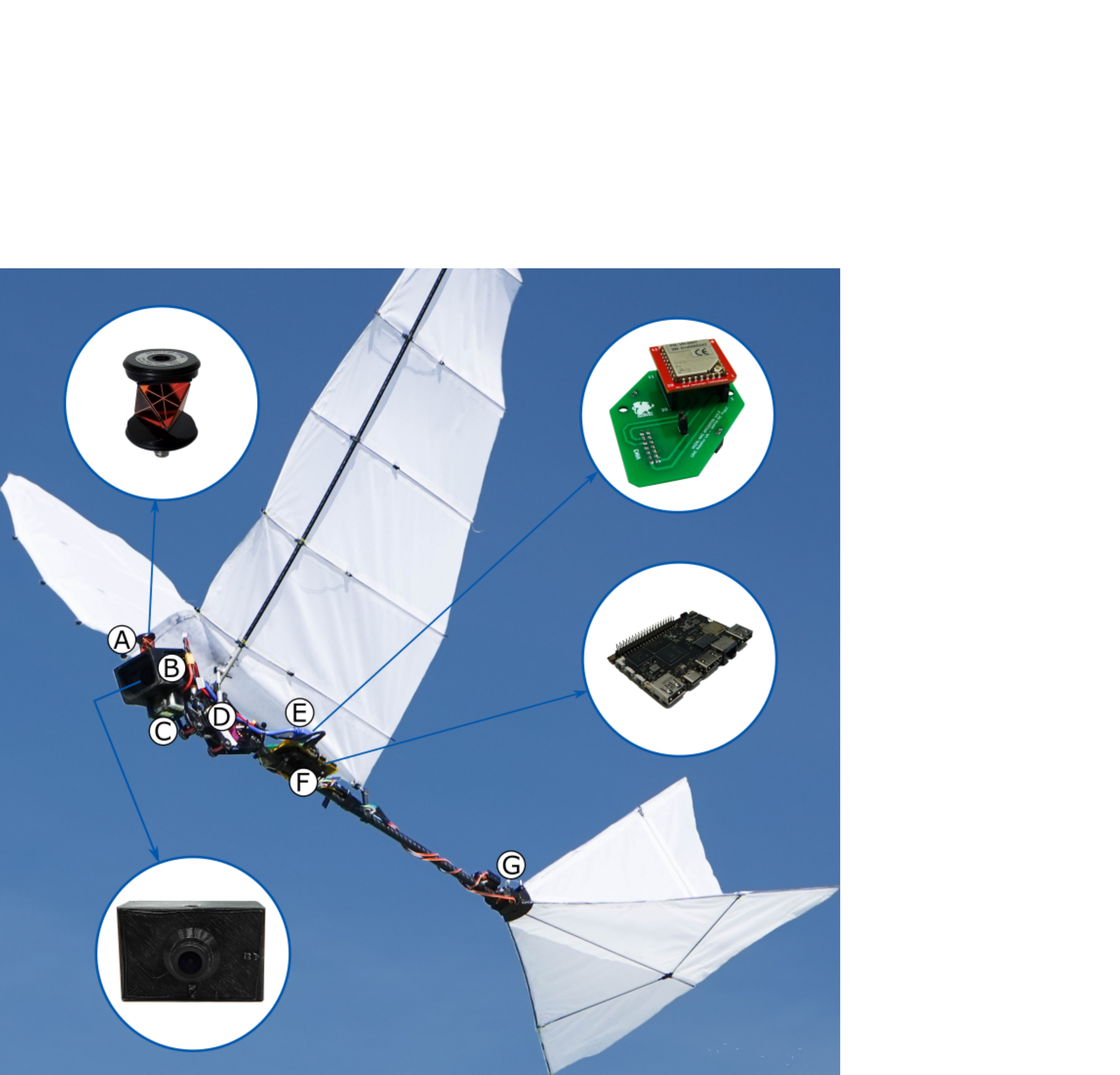}
    \caption{Design and parts distribution of GRIFFIN \textit{Eye-Bird}: Leica GRZ101 MiniPrism (A), DAVIS346 camera (B), LiPo battery (C), flapping mechanism (D), VectorNav VN-200 IMU (E), Khadas VIM3 board (F), and tail servos (G).}
    \label{fig:design}
    \end{center}
    \vspace{-4mm}       
\end{figure}

Figure \ref{fig:design} shows the ornithopter design and parts allocation. Placing sensors and batteries  at the head brings forward the CoG for better stability. Electronics were placed at the safest point, between the body tube and the wings. Although the exposed nut heads, bolts, and wires increase the total wetted surface of the aircraft and increases drag, this was preferred instead of covering the whole body with a fuselage, which would hamper electronics integration and would reduce  versatility.

Aerodynamic surfaces (wings and tail) are made of an ultra lightweight ripstop nylon fabric adhered to an optimized carbon fiber structure made of tubes and rods. This carbon fiber structure was designed as a tradeoff between the aerodynamic, elastic, and inertial properties of the wing. As a result, the wing is shaped for optimal low speed flapping flight aerodynamics, and capable of gliding with a low descent rate, which makes this ornithopter suitable for safe landing even in emergency cases, as opposed to multirotors. Wing fabric and structure achieve 82 g for a total surface of 0.44 m$^2$. Special emphasis was put into weight optimization over the wings. Lighter wings imply lower inertia forces during flapping, which become relevant at high speed flapping. Reducing wing inertia also reduces mechanical stresses over the flapping mechanism, and reduces the power demanded by the driving motor. Lower weight also gives more room available for additional payload.

The tail consists of a triangular horizontal stabilizer of 0.1 m$^2$ plus a triangular rudder of half the size,  providing longitudinal and directional stability and attitude control. Both surfaces are actuated with two concatenated servos, the first acting over the tail pitch, and the second, over the direction of the rudder. The tail pitch is used to control the aircraft nose pitch, and hence, the forward speed. The lack of roll actuations is mitigated by the roll stability that provides wing dihedral, and the directional stability and control provided by the rudder. The tail was trimmed for each weight distribution taking maximum glide ratio as criteria.

\subsection{Sensors}
\label{subsec:sensors}

The sensors on board GRIFFIN \textit{Eye-Bird} were selected considering their interest for addressing the flapping-wing robot perception challenges and the platform strict weight and size constraints. Data acquisition and collection was done with low-weight Khadas VIM3 board that equips a 6-core ARM CPU, an USB-3.0 interface, and a 16GB eMMC storage unit. The Khadas mounts Ubuntu 18.04 with ROS Melodic, and stores the collected data in the ROS bag format. 

A significant effort was devoted to reduce the weight of the components on board the ornithopter. The main characteristics of onboard sensors are shown in Table \ref{tab:my_label}, including their weights after the adaptation to integrate them on the ornithopter. The total weight of all sensors, electronics, and batteries, including the required cables, was lower than 250 g, near to the maximum payload capacity for the ornithopter with the desired maneuverability.

\begin{figure}[b!]
    \vspace{-2mm}   
    \begin{center}
    \includegraphics[width=0.85\linewidth, height=3.3cm]{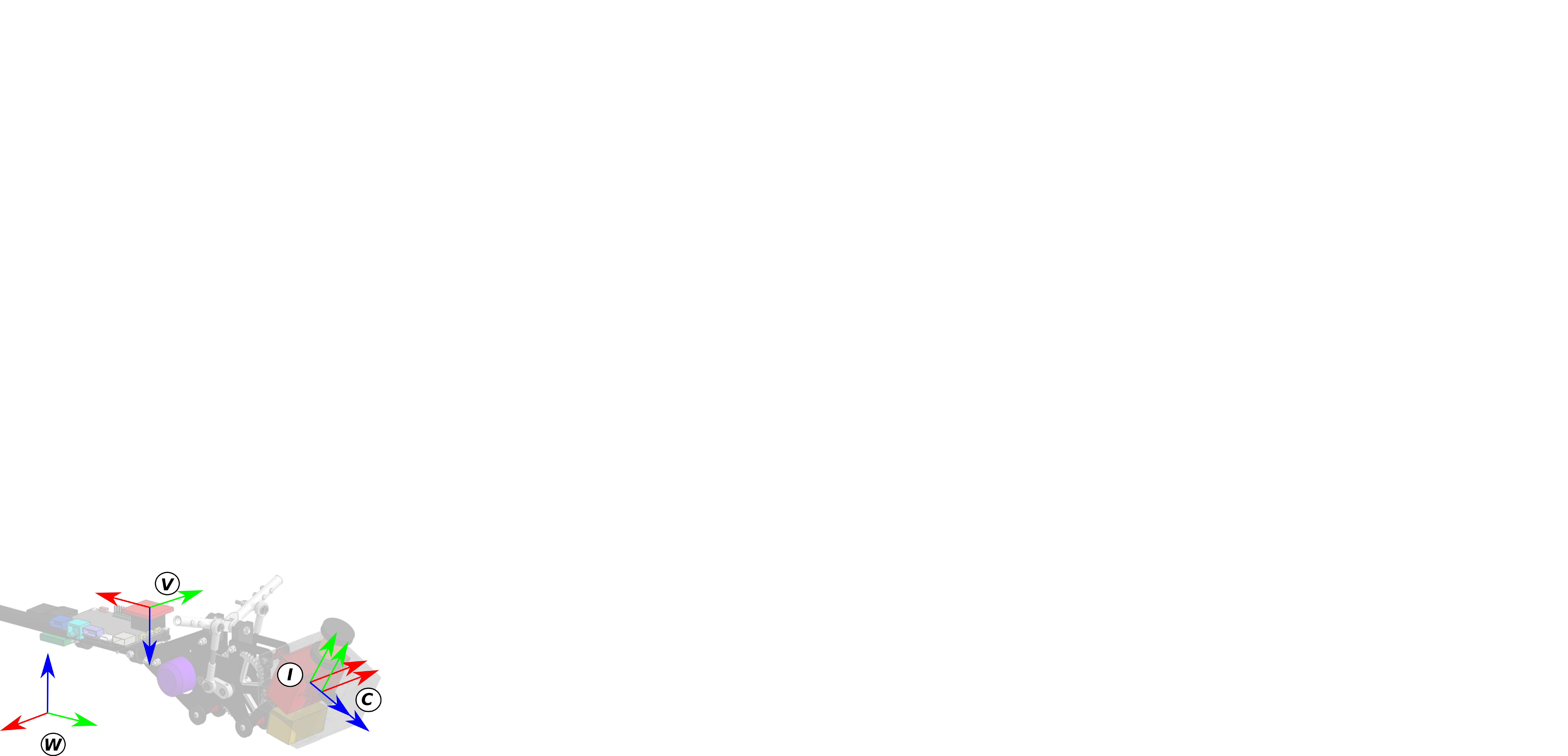}
    \caption{The GRIFFIN \textit{\textit{Eye-Bird}} ornithopter with the reference frames of each sensor: VectorNav (V), DAVIS346 APS and DVS (C), DAVIS346 IMU (I), and world reference frame (W) either from the TotalStation or the OptiTrack. The DAVIS346 was mounted upside down, affecting frames (C) and (I). The TotalStation prism is located at coordinate (0, 37.2, 0) mm at frame (I).}
    \label{fig:tf}
    \end{center}
\end{figure}

\begin{table}[]
    \begin{tabular}{c c c}
    \specialrule{.1em}{.05em}{.05em} 
        Sensor & Characteristics & Weight \\ \hline
        \multirow{4}{*}{DAVIS346} & DVS: 346x260 up to 12 MHz &   \\
        & APS: 346x260@40 Hz & With custom\\
        & FOV: 68 vert., 83 horiz. & case \& lens: 57 g\\
        & IMU: MPU 9250@1 kHz &\\ \hline
        \multirow{4}{*}{VN-200} & IMU readings @80 Hz &  \\ 
         & Gyroscope, accelerometer, & With adaptation\\
         & and magnetometer available  &  board: 21 g\\ 
         & GPS and barometer disabled  & \\ \hline \hline
        \multirow{2}{*}{\begin{tabular}[c]{@{}c@{}}Leica MS50\\TotalStation\end{tabular}} & Cent.-level tracking@10 Hz & \multirow{2}{*}{Prism: 30 g} \\ 
         & GRZ101 MiniPrism on the bird &  \\ \hline
        \multirow{2}{*}{OptiTrack} & Cent.-level tracking@100 Hz & LEDs and \\ 
        & 5x850nm LEDs on the bird  & cables: 5 g \\
    \specialrule{.1em}{.05em}{.05em} 
    \end{tabular}
    \caption{Onboard sensors and ground truth instruments.}
    \label{tab:my_label}
    \vspace{-0.4cm}
\end{table}
The main sensors mounted on GRIFFIN \textit{Eye-Bird} are an iniVation DAVIS346 and a VectorNav VN-200. The DAVIS346 embeds three different sensors: (1) a 346x260 dynamic vision sensor (DVS) that outputs timestamped and polarized events at a maximum rate of 12 MHz and with a temporal resolution of 1 $\mu$s; (2) a 346x260 active pixel sensor (APS) that is coincident with the DVS and outputs grayscale images at 40 Hz; and (3) a MPU 9250 IMU that delivers measurements at 1 kHz. To reduce its weight and fulfill the GRIFFIN \textit{Eye-Bird} strict payload and weight distribution requirements, the lens and metallic case of the DAVIS346 have been removed and substituted by lighter components. Two different lenses, one for outdoors and one for indoors were used, the latter with an IR cut-off filter to cope with the OptiTrack IR emitters. Each had a weight of 5 g, a focal distance of 3.6 mm and a Field of View of 83$^\circ$ horizontal and 68$^\circ$ vertical. The case was replaced with a PLA (Polylactic Acid) 3D printed encapsulation. The total weight of the adapted DAVIS346 was 52 g, less than a third of its original weight, which was $\sim~170$ g (original lens included). The DAVIS346 was mounted at the head of the ornithopter with a pitch angle of 30$^\circ$, in a protective lightweight soft case --made by 3D printing using flexible TPU95 filament-- that acts as a shock absorber in case of a frontal collision. To facilitate installation in the ornithopter head, the DAVIS346 was mounted upside down as seen in Figure \ref{fig:tf}, which shows the reference frames of each sensor. The VectorNav VN-200 includes a high-end IMU. It provides several times smaller noise density and in-run bias stability than the IMU embedded in the DAVIS346. It was installed as close as possible to the ornithopter center of gravity. The measurements from the VN-200 are accessed trough UART and published in ROS at 80 Hz using a custom package\footnote{{\url{https://github.com/grvcPerception/vn\_ros\_integration}}}. The VectorNav can also mount a GPS, but was not installed due to the flight and maneuverability weight constraints. Instead, the position ground truth was obtained with a motion capture system (indoor scenarios) and a Leica Nova MS50 TotalStation laser tracker (outdoor scenarios), which provide significantly lower localization errors than GPS.

The MS50 TotalStation is able to follow a moving reflective target and provide precise range and bearing measurements of it, which can be translated to 3D cartesian coordinates. The mounted target is a GRZ101 360$^\circ$ MiniPrism, which was installed at the head of the ornithopter. 
To prevent occlusions, the TotalStation was placed at an obstacle-free location and as far as possible from the flight area to increase the TotalStation Field of View. Further, to prevent occlusions caused by the ornithopter body, the flights were arranged such that the prism faced the TotalStation during the trajectory. The strong ornithopter motions while flapping could occasionally cause temporal tracking failures that only affected to a fraction of the recorded data, enabling a reliable ground truth position reconstruction in outdoor
flights. For indoor
flights,
an OptiTrack motion capture system with 28 cameras was used to provide millimeter accuracy ground truth position and orientation at a rate of 100 Hz. The ornithopter was tracked using five infrared (850 nm) LEDs installed at its main frame. The OptiTrack designated body frame was located at the position of the TotalStation prism. 

\section{Data Format and Calibration}
\label{subsec:dataset}

All sensor measurements were recorded as ROS bag files using timestamped messages. The format of each sensor measurement is defined by the standard ROS message library. Besides, the events from the DAVIS346 DVS use the format defined in \cite{mueggler2017event}, in which each event $e$ is represented as $e = (t_s,u,v,p)$, where $t_s$ is the time the event was triggered, $(u,v)$ are the event pixel coordinates, and $p$ is the event polarity either positive or negative. The images from the DAVIS346 APS include the pixel raw data, image resolution, and encoding. Further, the measurements from the VectorNav and DAVIS346 IMUs include referenced  orientation, angular velocity, and linear acceleration. The measurements from the TotalStation and the Optitrack are formatted as timestamped poses (without orientation in the case of the TotalStation). Each measurement is referenced w.r.t. its sensor frame, see Figure \ref{fig:tf}. The world reference frame is taken as the frames of the TotalStation 
and the OptiTrack.

A calibration dataset was obtained before each set of flights. The raw dataset is provided so users can calibrate with their tool of preference. Each dataset includes images from the DAVIS346 APS, events from the DAVIS346 DVS, and measurements from the DAVIS346 IMU and the VectorNav. Additionally, a calibration file obtained with the Kalibr toolbox \cite{rehder2016extending} is provided for each calibration dataset. The camera intrinsic calibration is recomputed at each calibration experiment. The extrinsic camera-IMU calibration of both IMUs is fixed as the most consistent calibration obtained by Kalibr through all the datasets, since it is less likely to change between flights and more affected by noise in the calibration data. Also, a 2-hours dataset of the IMUs measurements with the bird still is provided so users can obtain IMUs characterizations, e.g., using the Allan variance method\footnote{\url{https://github.com/rpng/kalibr\_allan}}. The calibrations were further validated by using them to execute a VIO (Visual-Inertial Odometry) algorithm in all the conducted flights, see Section \ref{sec:experimental}.

\section{The GRIFFIN Datasets}
\label{sec:experimental}

\begin{figure}[]
    \begin{center}
    \includegraphics[width=0.95\linewidth, height=5cm]{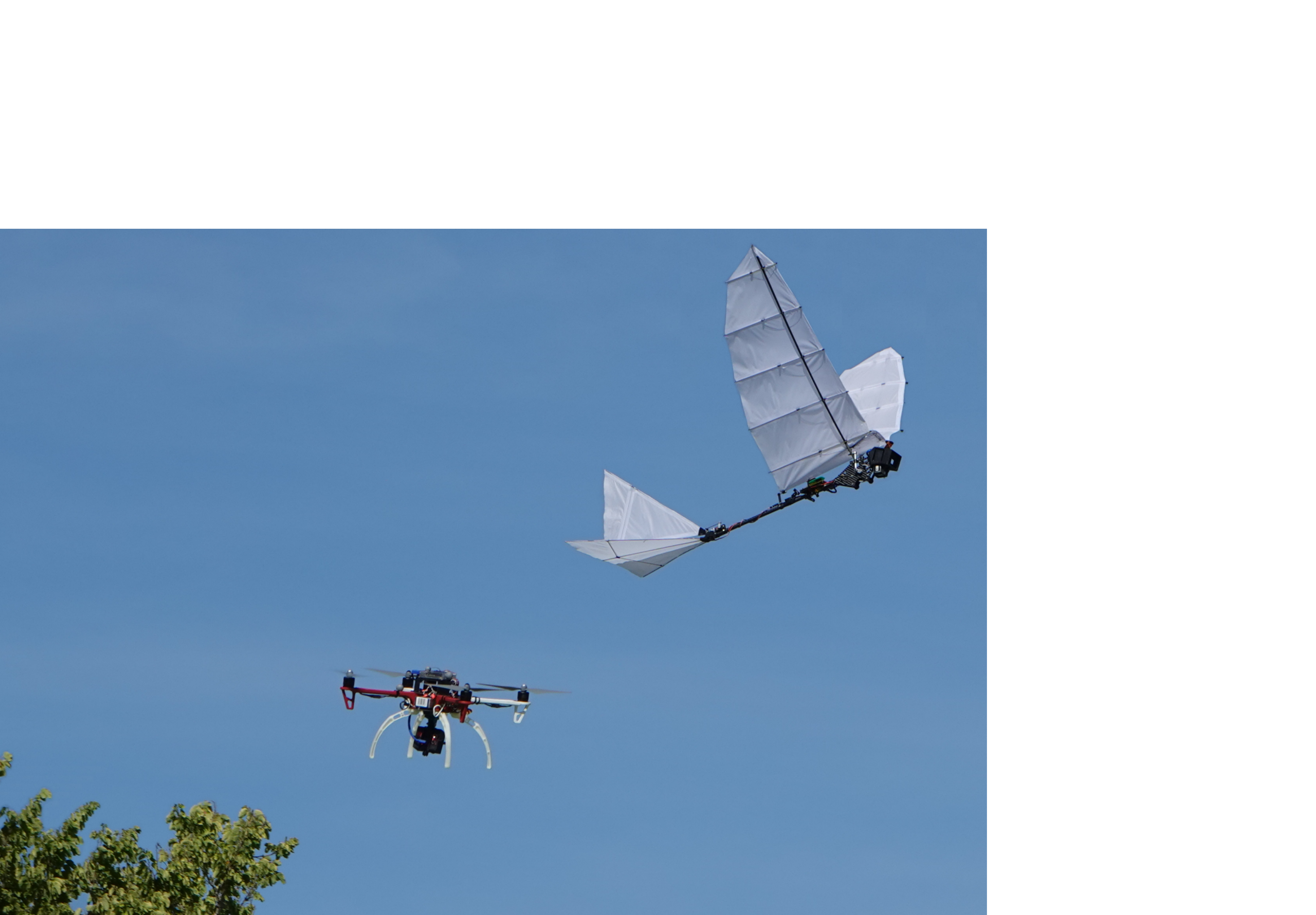}
    \caption{The DJI FlameWheel F450 platform used to compare the effect of wing flapping on onboard sensors.}
    \label{fig:multirotor}
    \end{center}
\vspace{-0.5cm}
\end{figure}

First, to better understand the effect of wing flapping on the onboard sensors, we performed a set of 
flights where the trajectories executed by the ornithopter were imitated by a multirotor UAV. The 
flights were repeated until the multi-rotor trajectories were similar to those performed by the ornithopter. The multirotor platform was a DJI FlameWheel F450 frame with a \textit{PixRacer} autopilot equipped with a DAVIS346 (mounted with a 30$^o$ pitch rotation) and a \textit{Khadas VIM3} board for data logging using ROS, see Figure \ref{fig:multirotor}. 

\begin{figure}[t]
    \begin{center}
    \includegraphics[width=0.46\linewidth]{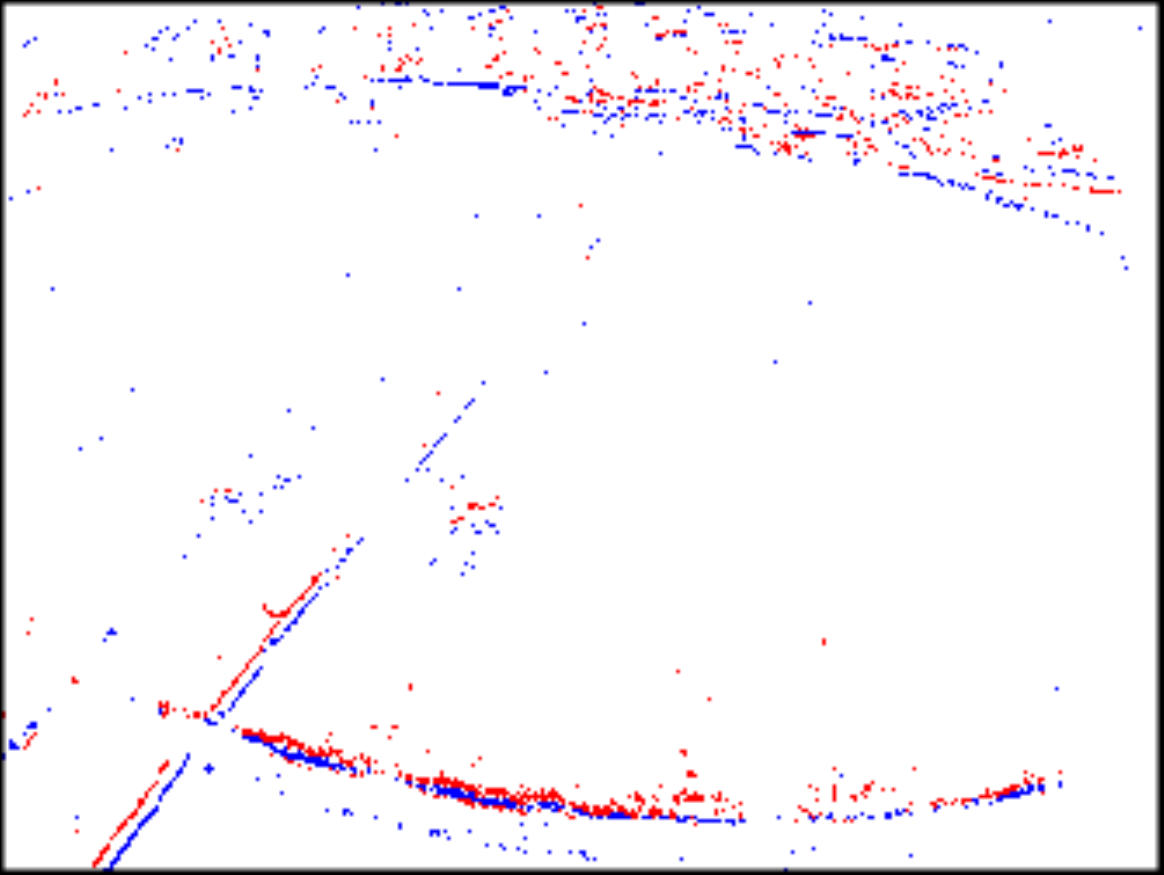}
    \includegraphics[width=0.46\linewidth]{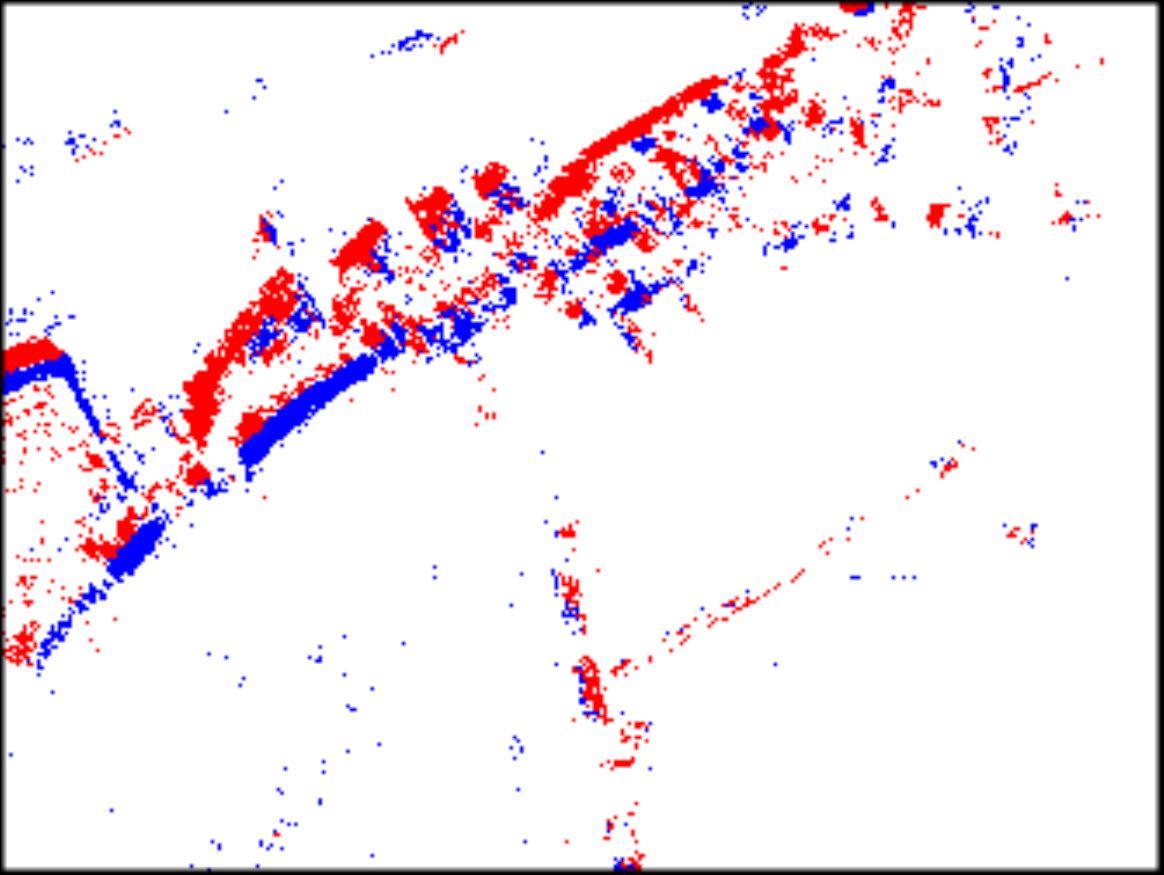}\\
    a) \hspace{4cm} b)
    \includegraphics[width=0.49\textwidth, height=3.6cm]{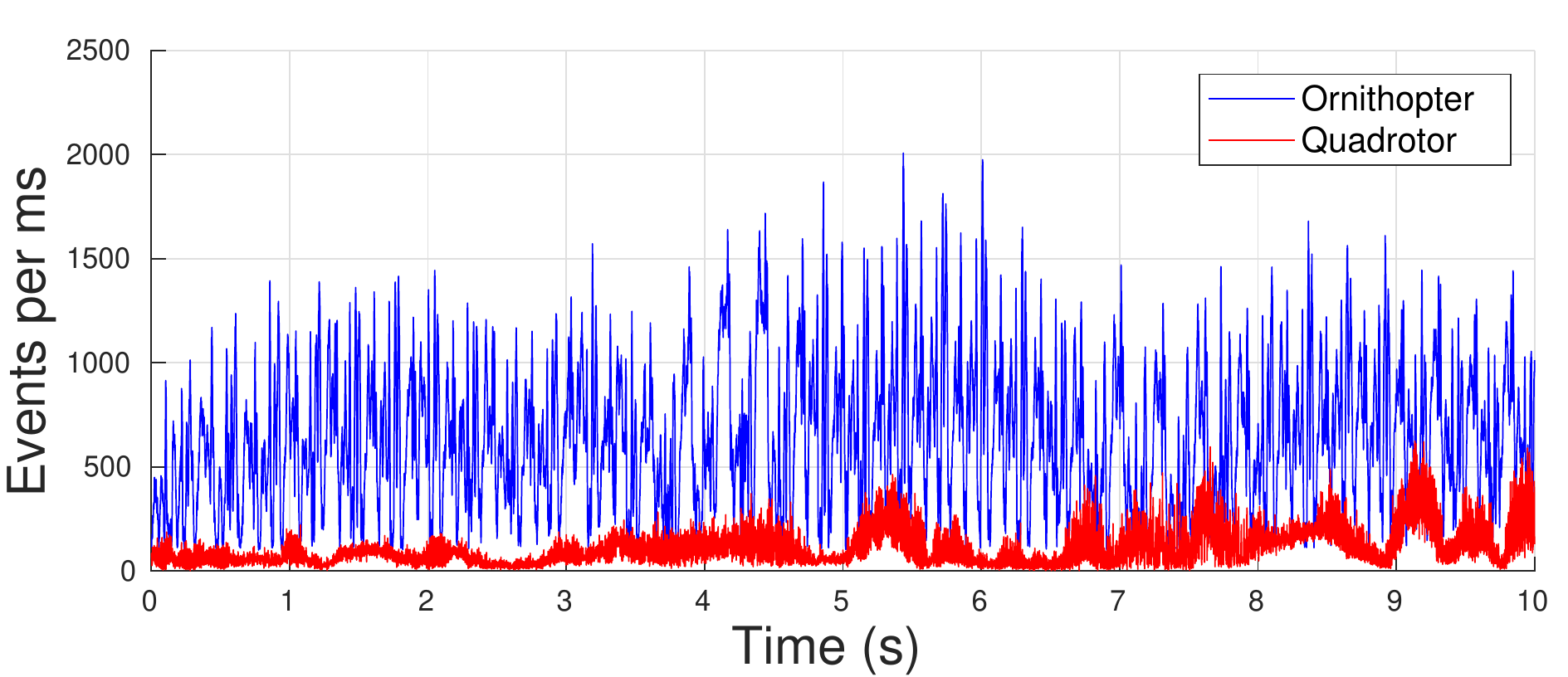}
    c)
    
    \caption{Comparison of number of events triggered per millisecond in ornithopter and multirotor flights describing a similar trajectory: a-b) event images accumulated at 40Hz on board the multirotor (a) and \textit{Eye-Bird} (b); and c) number of events accumulated every millisecond in both platforms.} 
    \label{fig:motion_blur}
    \end{center}
    \vspace{-0.5cm}
\end{figure}

As expected, the experimental results show that the ornithopter suffers from significantly stronger vibrations than the multirotor. On average the DAVIS346 IMU registered 3 times greater accelerations in \textit{Eye-Bird} than in the multirotor. These vibrations had direct impact on event generation: 6 times more events were triggered in \textit{Eye-Bird} than in the multirotor. Figure \ref{fig:motion_blur}-a,b shows the event images accumulated at 40 Hz gathered in both platforms when pointing to a similar part of the scenario. The ornithopter flight  generated a significantly greater number of events than the multirotor. Although this effect depends on the scenario and type of flight, these differences were consistently observed in all the flights performed. As an example, Figure \ref{fig:motion_blur}-c shows the number of events accumulated every millisecond while both platforms described a similar trajectory, confirming the intuition. Additionally, the abrupt changes in the ornithopter pitch angle during flapping often resulted in underexposed and overexposed visual images in outdoor scenarios, which imposes additional lighting robustness requirements to vision-based techniques suitable for ornithopters.

\begin{figure}[b!]
    \vspace{-2mm}   
    \begin{center}
    \includegraphics[trim={0 0 1.5cm 0.9cm},clip,width=0.495\textwidth,height=4.6cm]{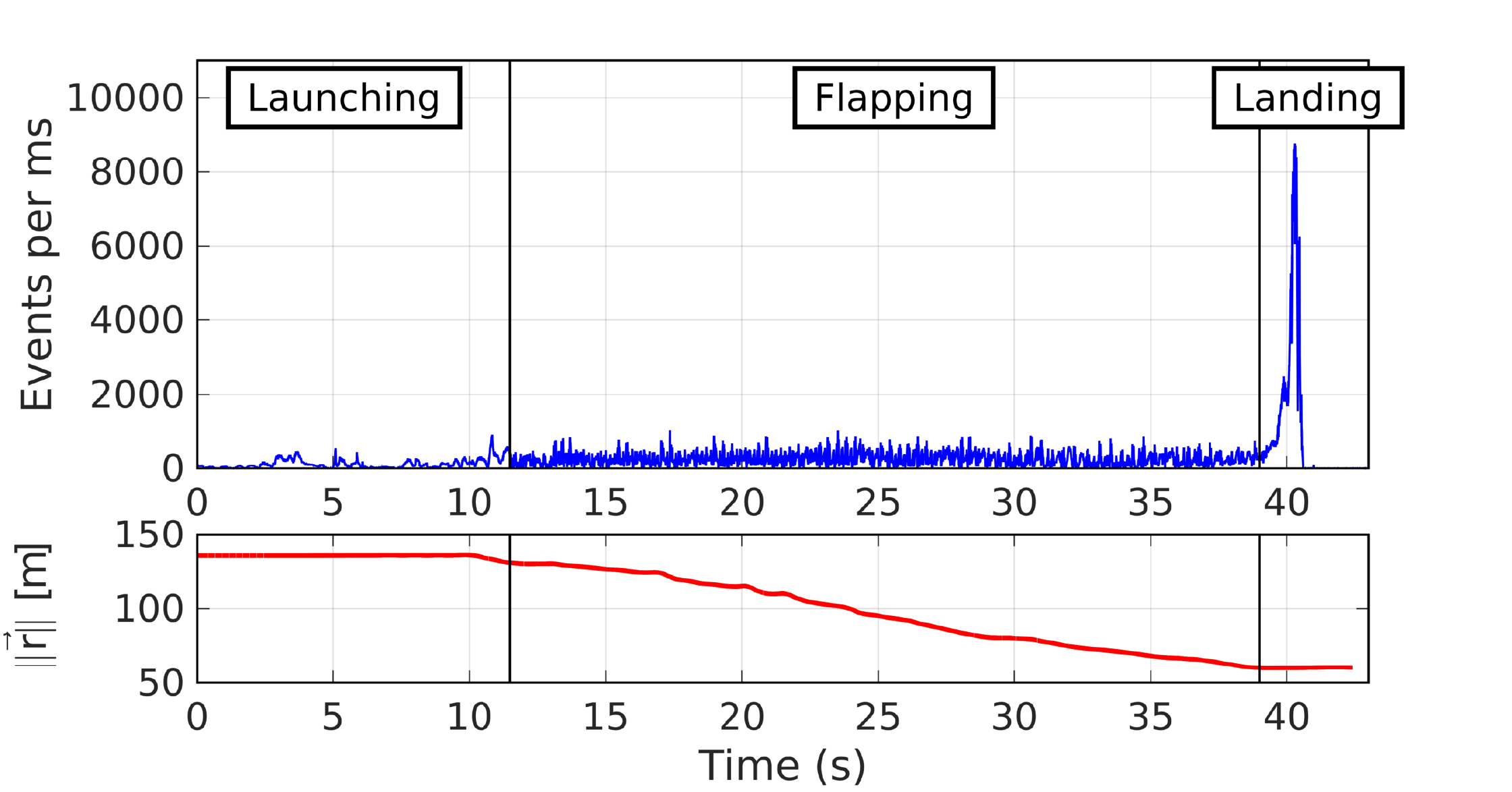}
    \caption{Number of events triggered per millisecond (top) and $\| \vec{r} \|$, module in world coordinates of the position vector of the TotalStation prism (bottom) in dataset \textit{Hills Base 3}.}
    \label{fig:continuidad}
    \end{center}
\vspace{-0.5cm}     
\end{figure}

The dataset collection consisted on recording data from onboard and external sensors during the ornithopter flight. %
The measurements from the onboard sensors were recorded in the Khadas VIM3, while those from the TotalStation and OptiTrack were recorded in an external computer. %
The clocks in the Khadas VIM3 and in the external computer were synchronized using Network Time Protocol (NTP) \cite{mills1991internet}.

\begin{figure*}[ht]
    \begin{center}
    \includegraphics[width=0.23\linewidth]{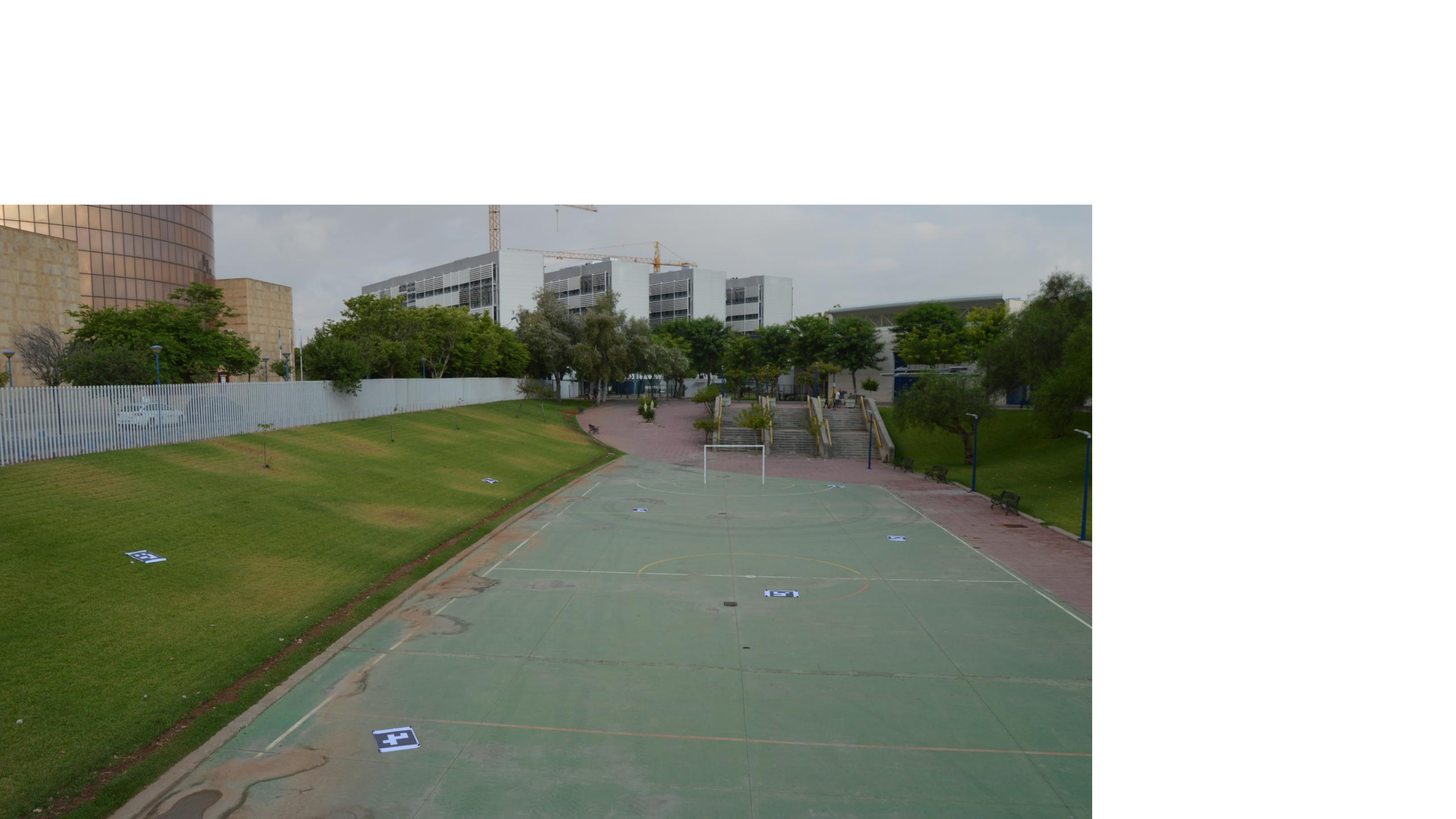}
    \includegraphics[width=0.23\linewidth]{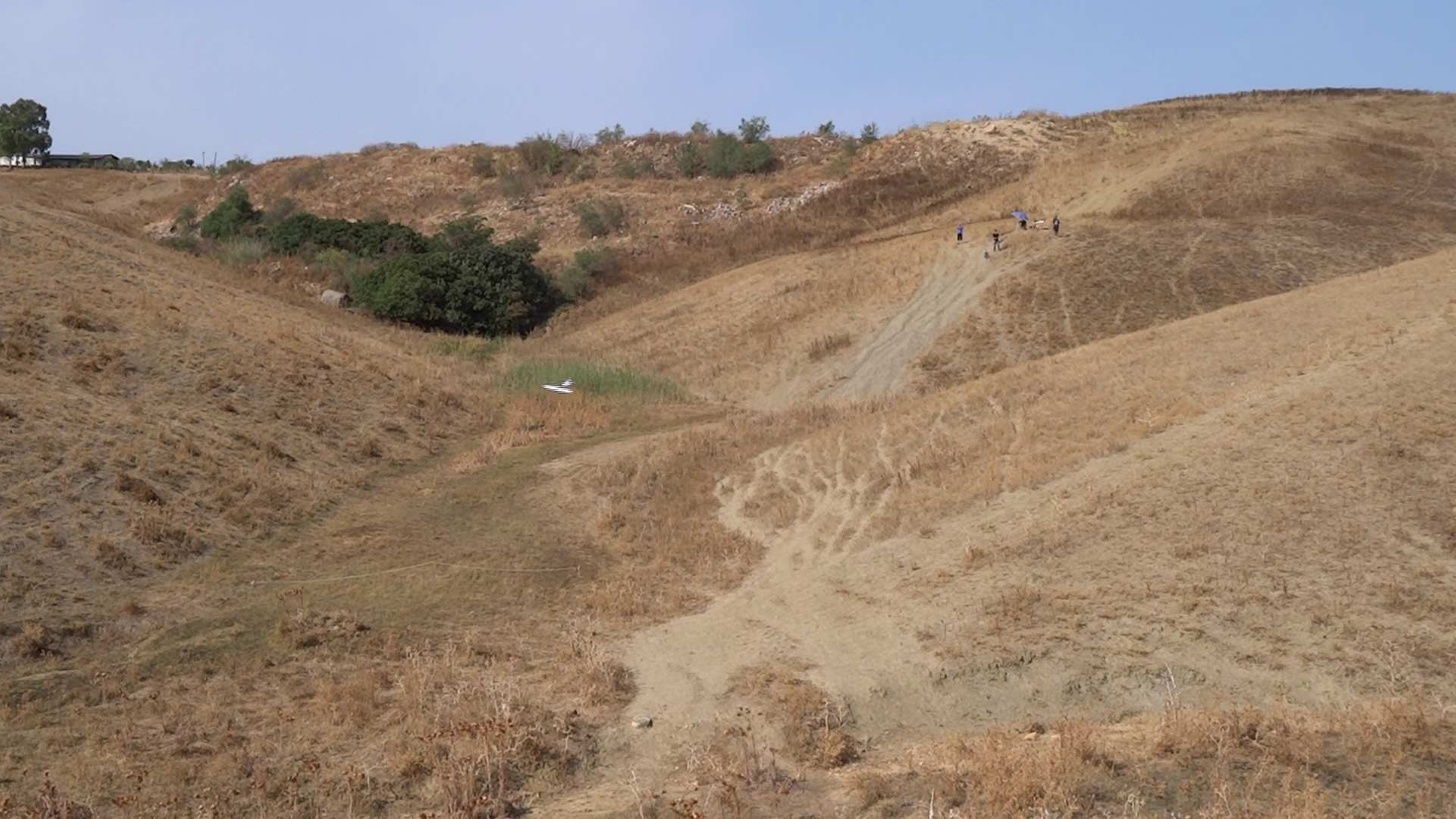}
    \includegraphics[width=0.23\linewidth]{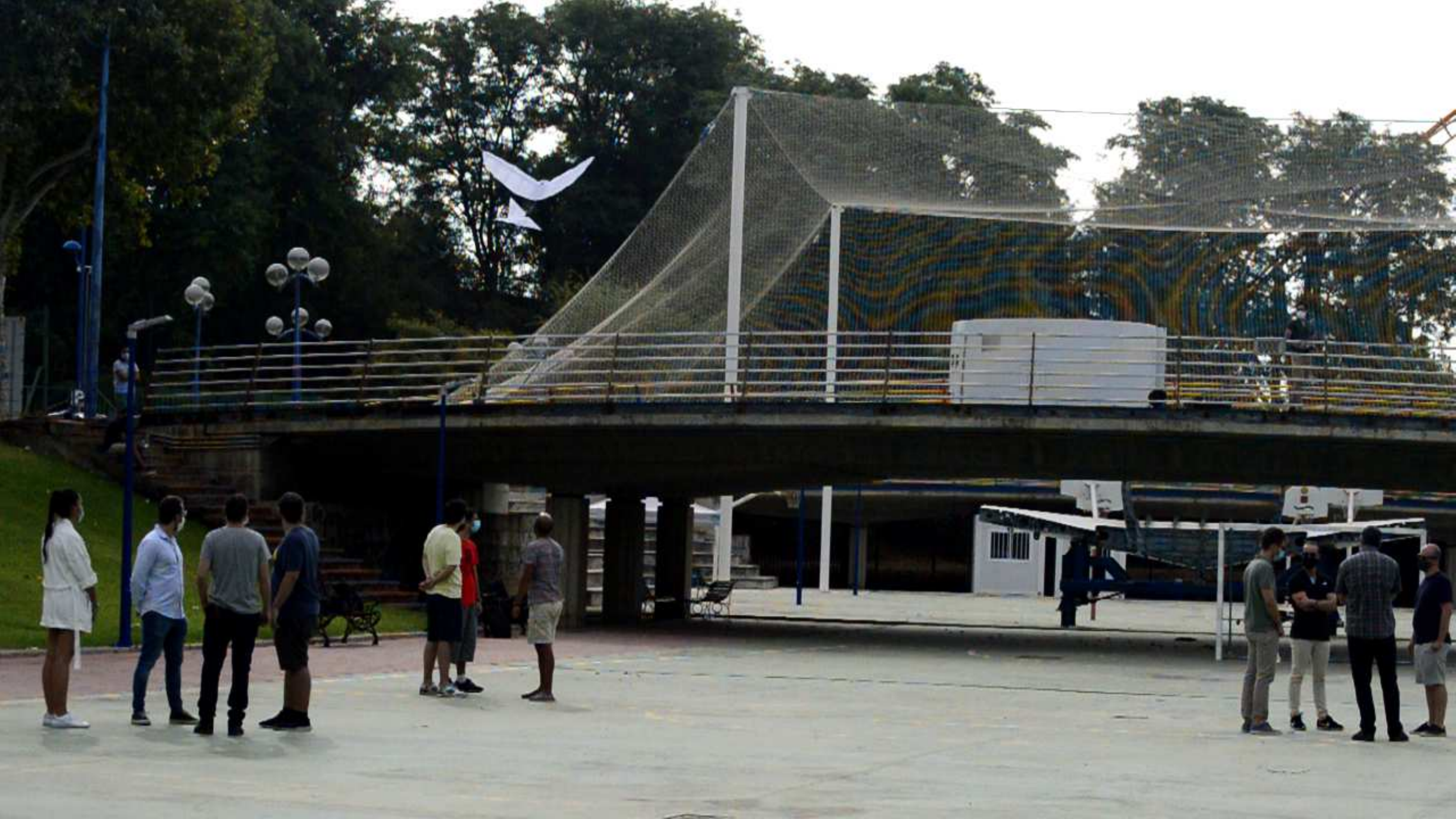}
    \includegraphics[width=0.23\linewidth]{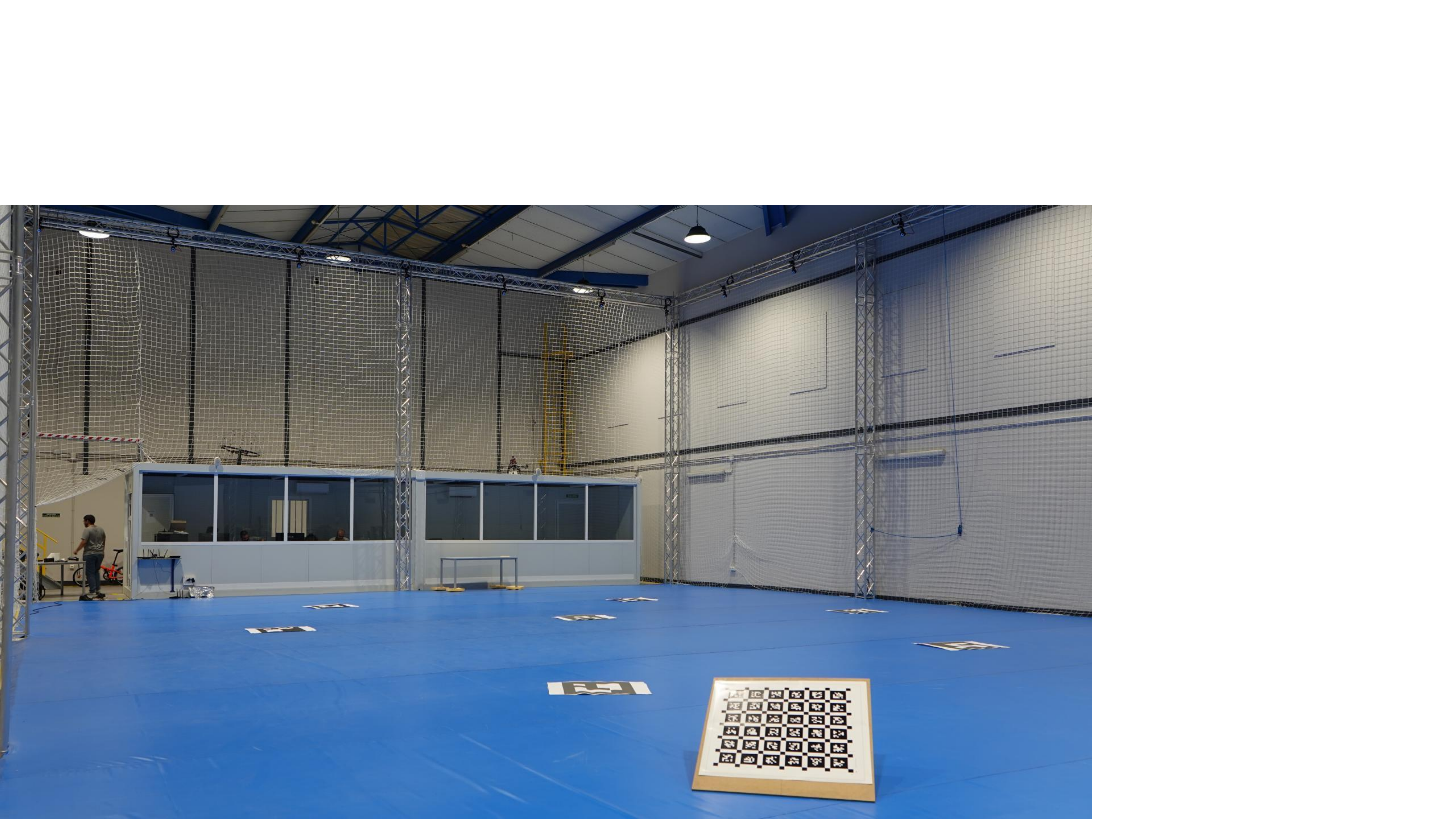}\\
    \vspace{0.7mm}    
    \includegraphics[width=0.23\linewidth]{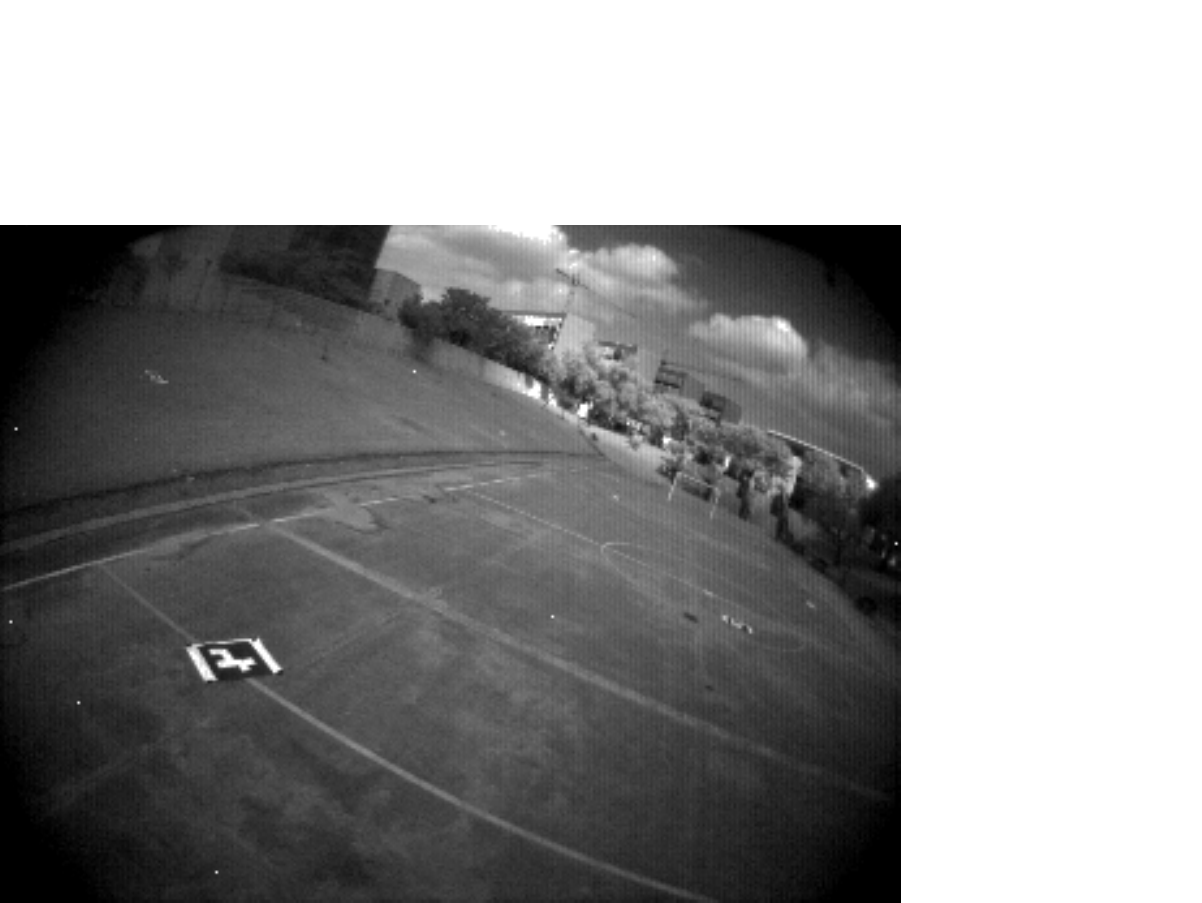}
    \includegraphics[width=0.23\linewidth]{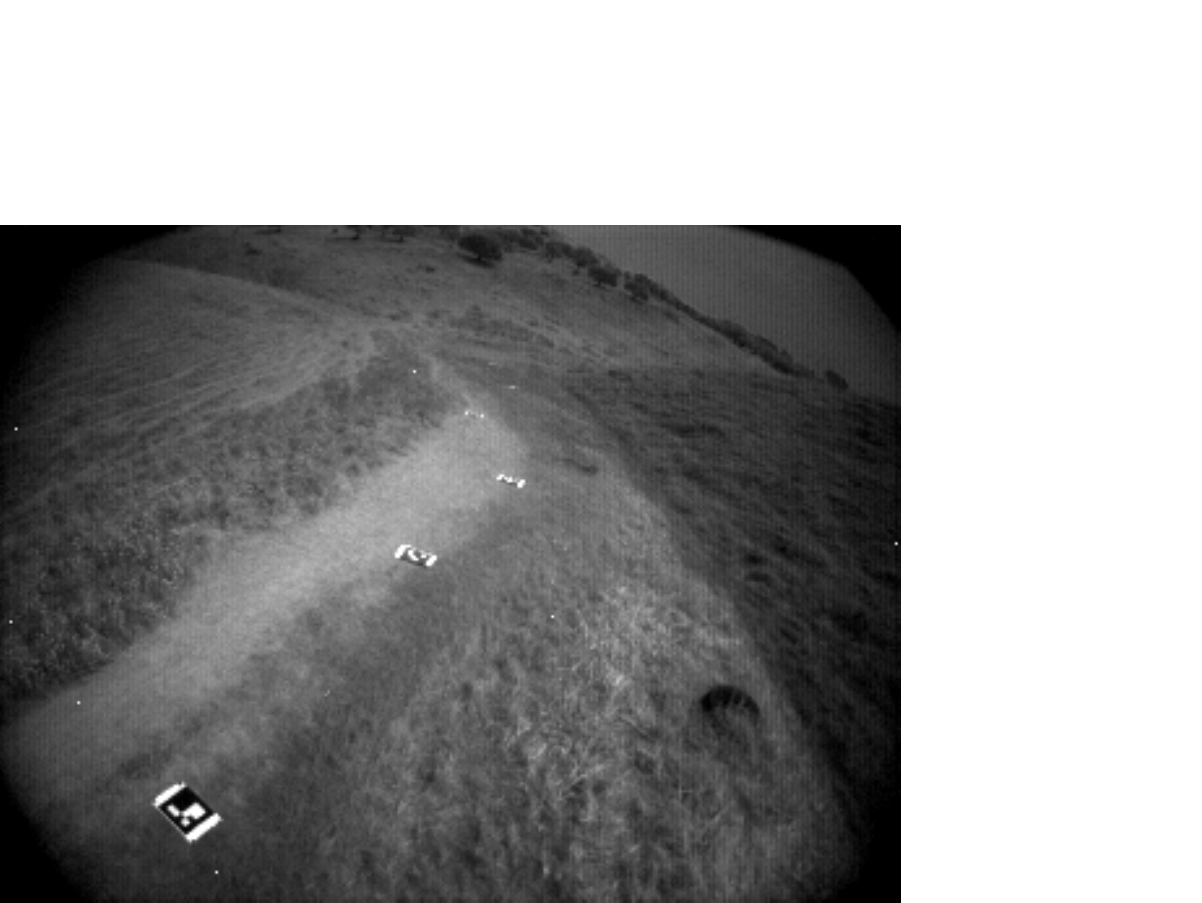}
    \includegraphics[width=0.23\linewidth]{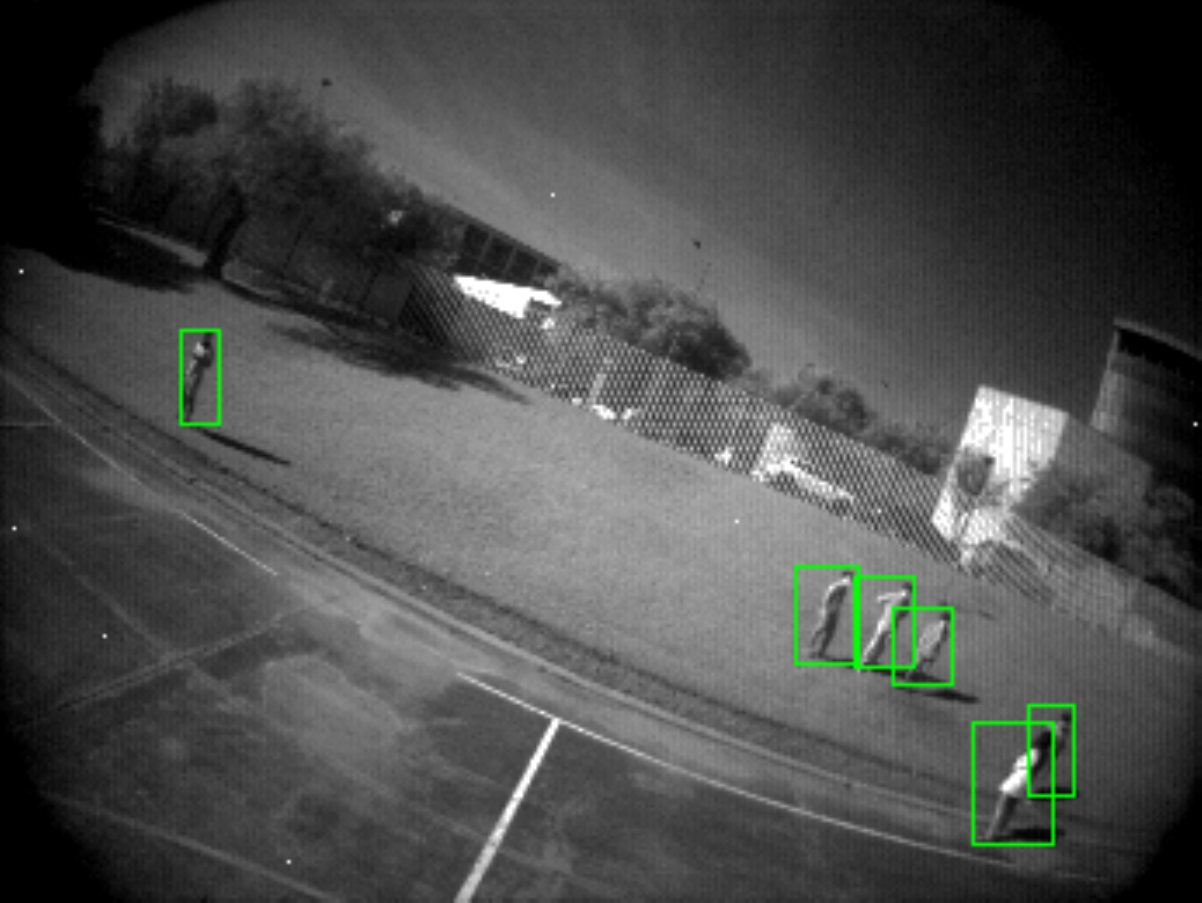}
    \includegraphics[width=0.23\linewidth]{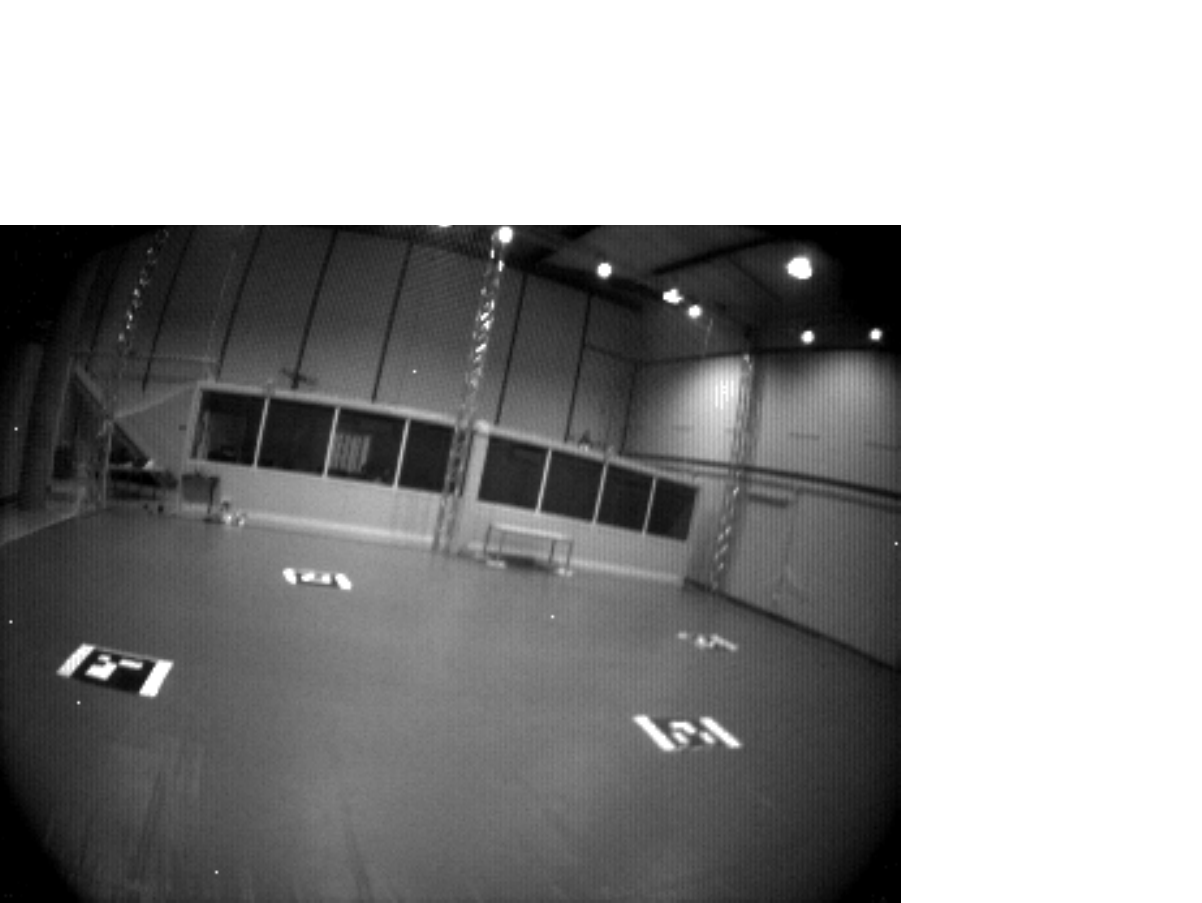}\\
    \vspace{0.7mm}    
    \includegraphics[width=0.23\linewidth]{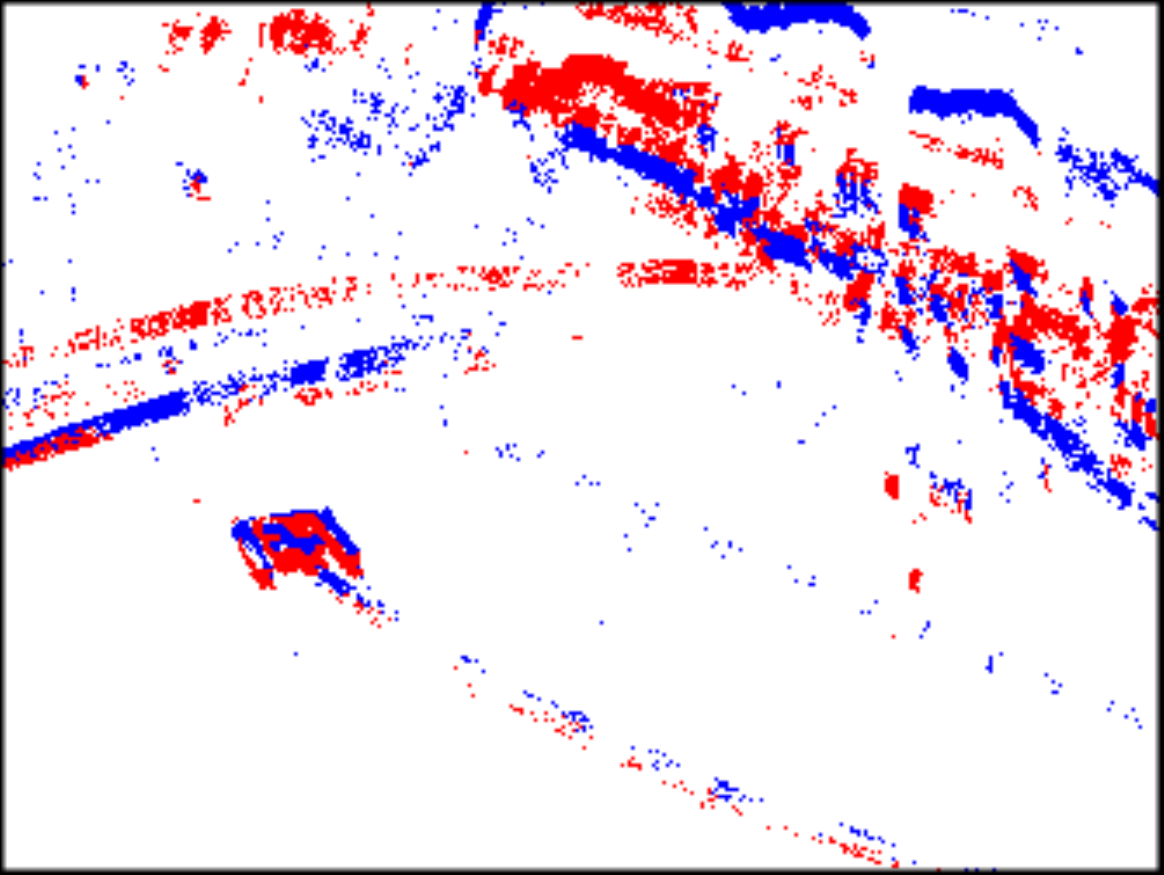}
    \includegraphics[width=0.23\linewidth]{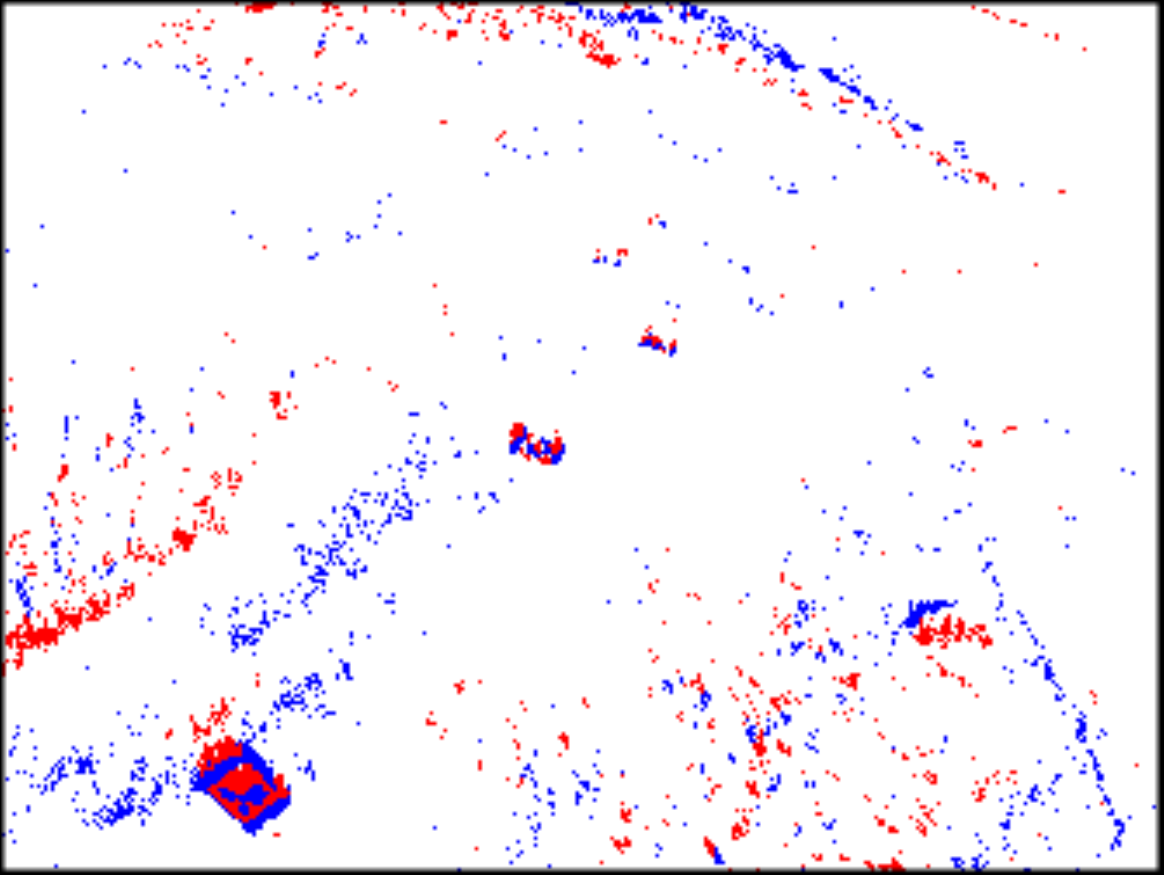}
    \includegraphics[width=0.23\linewidth]{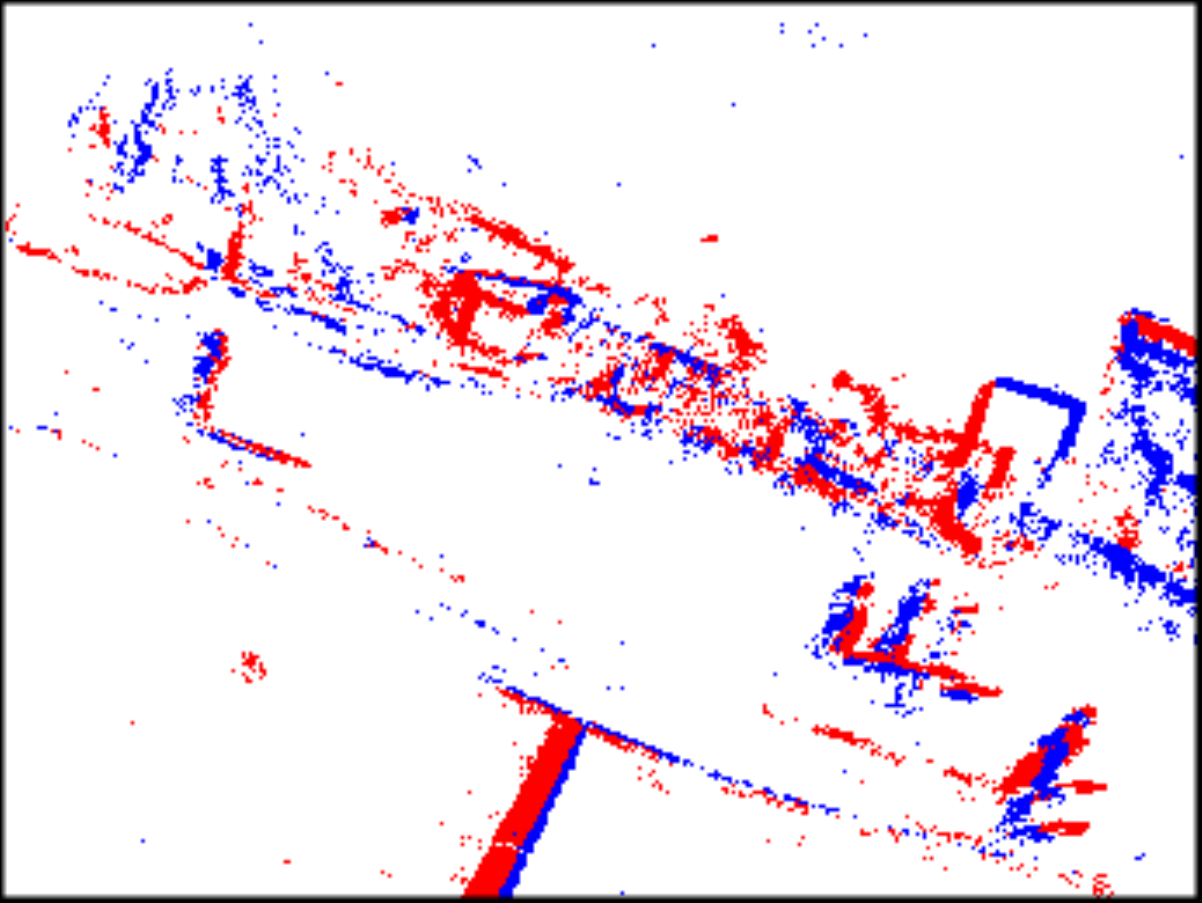}
    \includegraphics[width=0.23\linewidth]{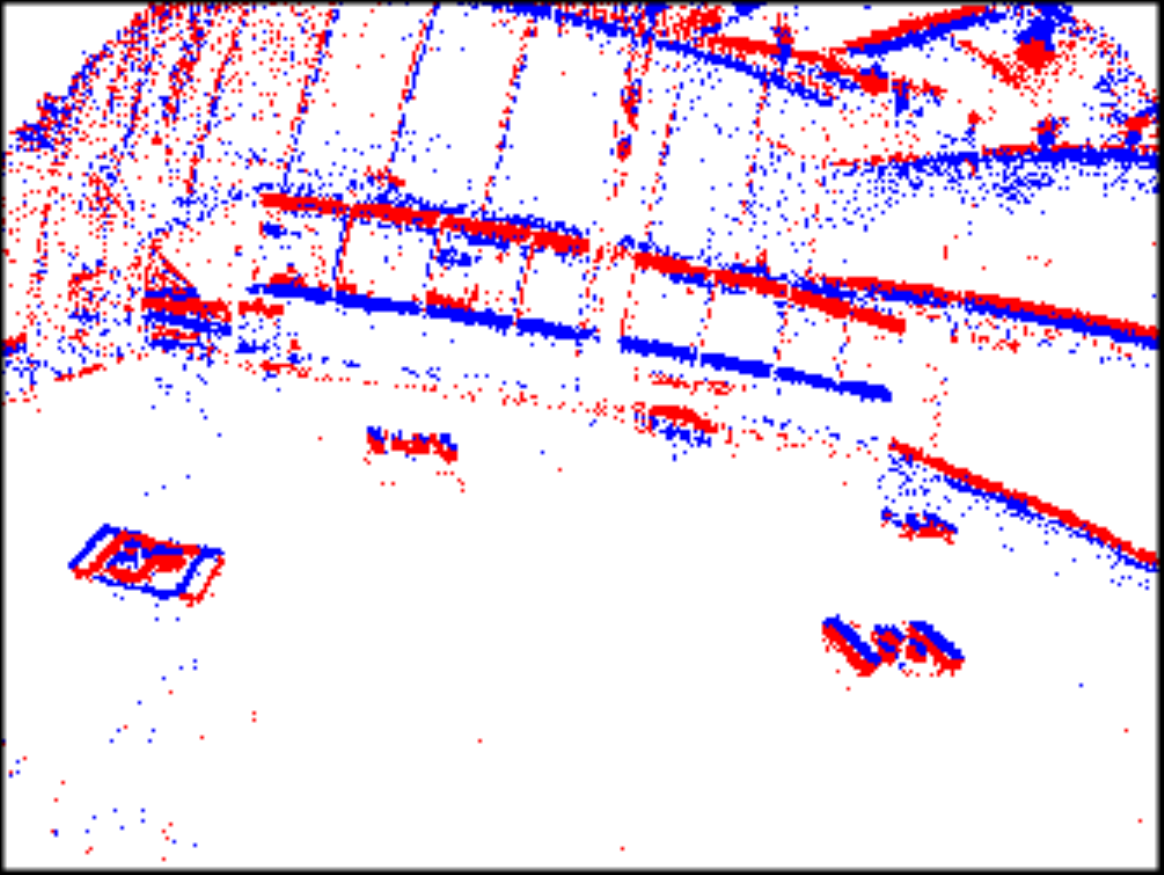}\\
    a) \hspace{3.6cm}  b)  \hspace{3.6cm}  c)  \hspace{3.6cm}  d)
    \caption{Pictures from some 
    flights of the dataset: a) \emph{Soccer}, b) \emph{Hills}, c) \emph{People} 
    flight in the \emph{Soccer} scenario, d) \emph{Testbed}.} 
    \label{fig:scenarios}
    \end{center}
    \vspace{-2mm}       
\end{figure*}

Data recording (of onboard and external sensors) was initialized before each flight. Just before being launched, the ornithopter was oriented such that the DAVIS346 pointed towards a referenced calibration grid in order to add visual reference and features. Next, the ornithopter was smoothly launched towards the flight area. Once in the air, the ornithopter was guided towards the landing area while performing different types of trajectories. 
Figure \ref{fig:continuidad}-top shows the number of events generated per millisecond during one of the dataset flights (\textit{Hills Base 3}). The three flight stages --launching, flapping, and landing-- can be inferred from the graph, since the event generation is affected by different dynamics. Figure \ref{fig:continuidad}-bottom shows $\| \vec{r} \|$, the module in world coordinates (W) of the position vector of the TotalStation prism throughout that flight. Temporal synchronization is noticed by comparing both graphs (e.g., the maximum number of events per millisecond matches the ornithopter landing). The ornithopter flights had different durations depending on the maneuvers performed and the type of dataset. Three types of datasets were recorded:

\begin{table*}[t!]
\centering
\scalebox{0.99}{
\begin{tabular}{c c c c c c c c c c} 
\specialrule{.1em}{.05em}{.05em}
Scenario &  Dataset & $t$ (s) & $d$ (m)  &  $\left\| \bar{v} \right\|$ (m/s) & $\left\| \bar{w} \right\|$ (rad/s) & Max EPms (events/ms) & $t_{RMSE}$ (m) & Ground Truth & Annotation \\ 
\hline 
\multirow{9}{*}{Testbed} & Base1 & 27.99 & \phantom{0}96.12 & 3.16 & 0.29 & 3575 & 0.68 & OptiTrack & No\\ 
           &  Base2 & 18.99 & \phantom{0}76.26 & 3.91 & 0.28 & 8171 & 0.59 & OptiTrack & No\\ 
           &  Base3 & 16.99 & \phantom{0}57.61 & 3.14 & 0.25 & 4094 & 1.03 & OptiTrack & No\\
           &  ArUco1 & 20.98 & \phantom{0}61.37 & 2.81 & 0.21 & 4295 & 0.41 & OptiTrack & No\\
           &  ArUco2 & 24.99 & \phantom{0}94.10 & 3.42 & 0.27 & 5073 & 0.55 & OptiTrack& No\\
           &  ArUco3 & 26.98 & 108.38 & 3.86 & 0.29 & 8726 & 0.46 & OptiTrack & No\\ 
           &  ArUco4 & \phantom{0}4.80 & \phantom{0}12.17 & 2.57 & 0.14 & 8631 & 0.06 & OptiTrack & No\\
           &  ArUco5 & \phantom{0}3.97 & \phantom{0}11.08 & 2.83 & 0.13 & 8616 & 0.13 & OptiTrack & No\\
           &  ArUco6 & \phantom{0}3.98 & \phantom{0}11.98 & 2.99 & 0.12 & 8667 & 0.21 & OptiTrack & No\\ 
\hline
\multirow{5}{*}{Hills} & Base1 & 39.21 & 107.76 & 2.77 & -- & 8704 & 1.20 & TotalStation & No\\ 
    & Base2  & 39.90 &  110.21 & 2.80  & -- & 8784 & 2.48 & TotalStation & No\\
    & Base3  & 42.91 & \phantom{0}98.75 & 2.52 & -- & 8772 & 2.46 & TotalStation & No\\
    & ArUco1  & 31.80 & \phantom{0}97.95  & 2.97 & -- & 8840 & 2.14 & TotalStation & No\\
    & ArUco2  & 27.21 & 114.93 & 4.31 & -- & 6615 & 6.46 & TotalStation & No\\
\hline
\multirow{7}{*}{\begin{tabular}[c]{@{}c@{}}Soccer\\\end{tabular}}  & Base1 & 24.89 & \phantom{0}54.65 & 2.22 & -- & 8610 & 0.83 & TotalStation & No\\ 
           & Base2  & 20.01 & \phantom{0}43.41 & 2.16 & -- & 8549 & 1.63 & TotalStation & No\\ 
           & Base3  & 27.90 & \phantom{0}54.42 & 1.92 & -- & 7166 & 1.91 & TotalStation & No\\ 
           & ArUco1  & 15.31 & \phantom{0}51.49 & 3.47 & -- & 7534 & 1.01 & TotalStation & No\\ 
           & ArUco2  & 53.02 & \phantom{0}19.50 & 2.79 & -- & 8757 & 1.12 & TotalStation & No\\ 
           & People1 & 79.98 & -- & -- & --  & 8737 & -- & None & Yes\\ 
           & People2 & 88.04 & -- & -- & --  & 8729 & -- & None & Yes\\ 
\hline

\specialrule{.1em}{.05em}{.05em} 
\end{tabular}
}
\caption{Summary of the provided datasets: $t$, flight time duration; $d$, total traversed distance; $\left\| \bar{v} \right\|$ and  $\left\| \bar{w} \right\|$, mean inertial linear and angular velocities; maximum number of events per millisecond (EPms) along each flight: $t_{RMSE}$, absolute translation root mean square error of the robot position estimated using ROVIO \cite{bloesch2017rovio}; and availability of annotated data. $\left\| \bar{w} \right\|$ is not provided in Hills and Soccer scenarios as TotalStation provides only position ground truth.}
\label{table:datasets}
\vspace{-2mm}
\end{table*}

\begin{itemize}
    \item In \emph{Base} datasets the ornithopter described
    agile
    trajectories until it reached a safety altitude for landing. There were no artificial landmarks in the scenario and the maneuver varied depending on the wind conditions. 
    \item In \emph{ArUco} datasets the ornithopter described smooth trajectories trying to flight over ArUco markers placed on the ground used to add ground truth references and features in the scene. 
    \item In \emph{People} datasets the ornithopter flies over people and objects with the aim
    of providing samples for object detectors. To cover larger areas, the ornithopter described longer trajectories without altitude limitation, and hence, these datasets were recorded without measurements from the TotalStation.
\end{itemize}

The datasets were recorded in two outdoor and one indoor environments, see Figure \ref{fig:scenarios}. The \emph{Soccer} outdoor scenario is a small soccer field surrounded by obstacles with different heights such as threes, benches, and fences. Its total area is $48\times54$ m. The ornithopter was launched from a small bridge structure in front the yard, while the TotalStation was located at an elevated spot located at $\sim 74$ m from the launching point. The \emph{Hills} outdoor scenario is a large obstacle-free irregular area with hills and slopes. It is a large scenario ($170\times100$ m), which enables longer flights without  obstacles or constrained flight space. The distance between the TotalStation and the launching spot was $\sim 135$ m. The \emph{Testbed} indoor scenario was a $15\times21\times8$~m room designed for testing ornithopters and equipped with an OptiTrack motion capture system. In the simplest indoor trajectories \textit{Eye-Bird} was controlled autonomously with a simplified method inspired in \cite{maldonado2020adaptive}, which used the OptiTrack measurements. These datasets can be of interest for developing onboard perception techniques for closing the low-level control loop. In the outdoor and the most complex indoor
flights \textit{Eye-Bird} was controlled manually by an expert pilot to better compensate for the wind disturbances and perform richer trajectories that exploit the ornithopter maneuverability and also evidence the challenges of flapping for perception.

Table \ref{table:datasets} lists the 21 provided datasets, 9 of them conducted in the \emph{Testbed} scenario, 7 in the \emph{Soccer} scenario, and 5 in \emph{Hills}. A total of 10 \emph{ArUco} and 9 \emph{Base} datasets are provided due to their interest for perception techniques in partially structured (\emph{ArUco}) and fully unstructured (\emph{Base}) environments. %
\emph{People} datasets were conducted in the \emph{Soccer} scenario. In these datasets annotated bounding boxes are provided to facilitate training for people detection algorithms. For safety, we did not %
record
\emph{People} 
datasets in the \emph{Testbed}. Also, in the \emph{Hills} scenario the robot flew at high altitudes, which hampered the correct identification of objects. Table \ref{table:datasets} also shows the maximum number of events generated per millisecond in each flight. The maximum event throughput of DAVIS346 is 12 Million Events Per Second, i.e. 12,000 events per millisecond. Hence, the number of events generated per millisecond in all flights were lower than the DAVIS346 throughput, i.e. no event was discarded in any dataset. For instance, Figure \ref{fig:continuidad}-top shows the events triggered during the \textit{Hills Base 3} flight. The number of generated events is always lower than the maximum event throughput of DAVIS346 even when the ornithopter impacts on the ground. In all conducted flights the maximum event rate takes place when the ornithopter impacts on the ground, and this value is at least 5 times bigger than the mean event rate during the rest of the flight.

\begin{figure*}[h!]
    \begin{center}
    \includegraphics[width=0.40\linewidth, height=5cm]{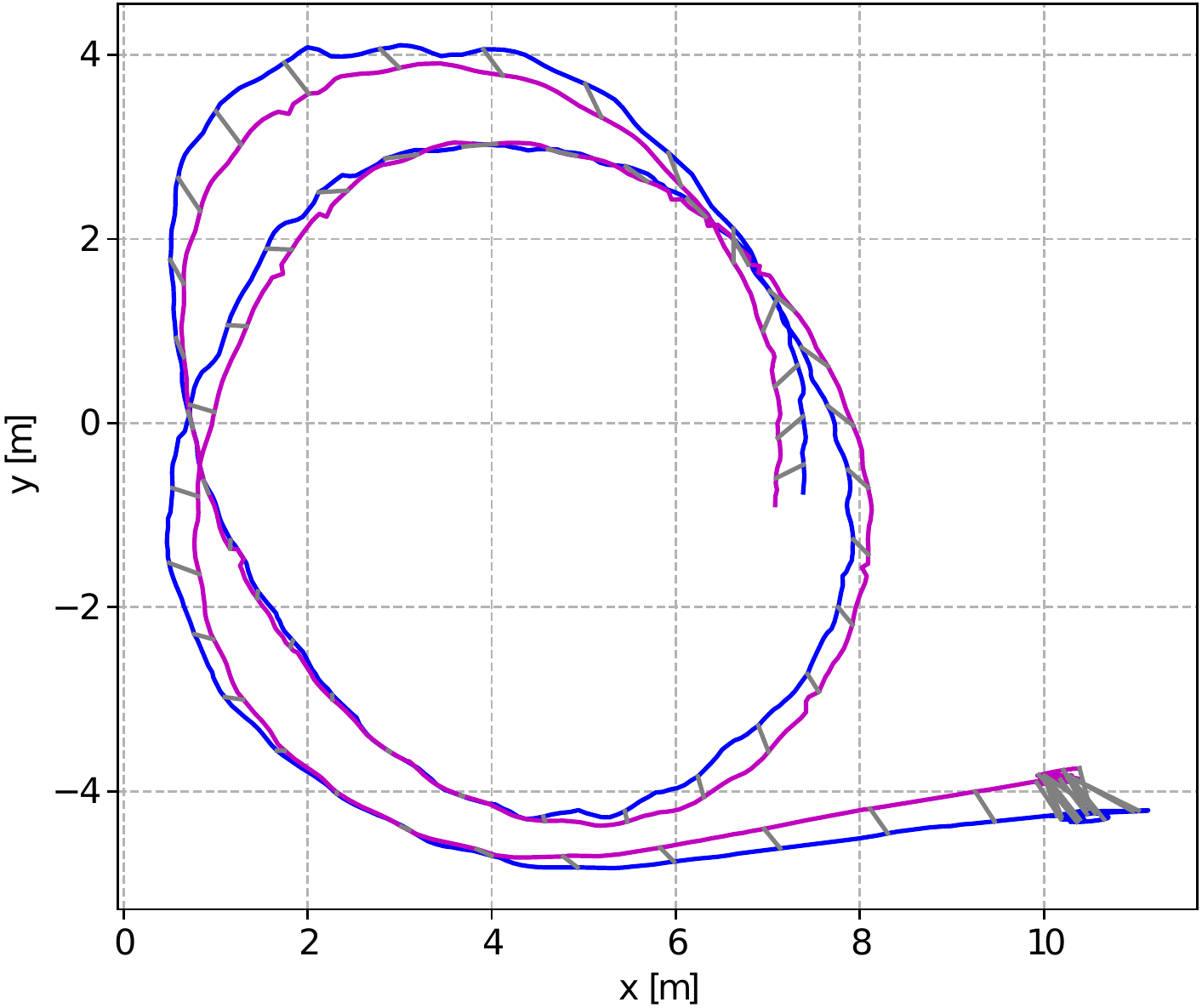}
    \hspace{0.10\linewidth}
    \includegraphics[width=0.40\linewidth, height=5cm]{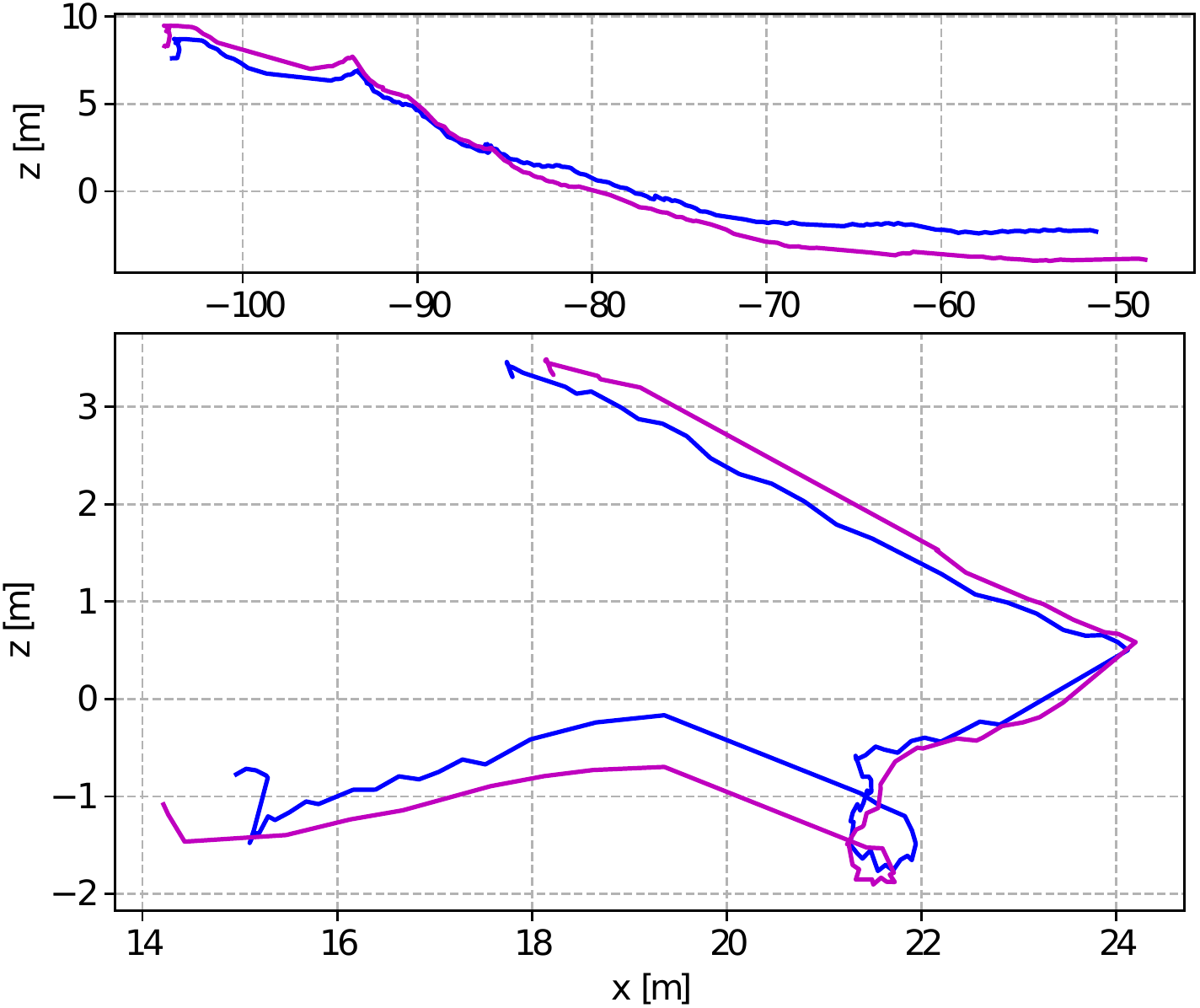}
    \caption{Comparison of robot trajectories estimated using ROVIO \cite{bloesch2017rovio} (in blue) versus ground truth (in magenta) in flights \textit{Testbed ArUco1} (Left), \textit{Hills ArUco1} (Right-up), and \textit{Soccer ArUco1} (Right-down).} 
    \label{fig:vios}
    \end{center}
\end{figure*}

To complement the dataset, a VIO algorithm was evaluated in all the flights and compared to the provided ground truth. This is valuable for confirming the usability and validity of the provided dataset, assessing the quality of the given calibration and ground truth, and providing a baseline result for comparison with other VIO methods. The chosen method was ROVIO \cite{bloesch2017rovio}, a robust and well-known method capable of providing estimates of the robot trajectory without intensive parameter tuning. Figure \ref{fig:vios} shows the ornithopter trajectory estimated with ROVIO versus the ground truth in one flight in each scenario. The absolute translation RMSE (root mean square error) of the VIO trajectory at each flight w.r.t. the position ground truth is provided in Table \ref{table:datasets}. The execution of ROVIO was cut before the landing, where no visual features are present. The absolute error provided by ROVIO in dataset \emph{Hills ArUco2} is caused by drift and confirms the limitation of existing techniques with flapping-wing flights.

\section{Conclusions and Future Work}
\label{sec:conclusion}

The development of perception techniques for flapping-wing robots faces a number of issues. First, the high mechanical vibrations and sudden motion of these platforms originates motion blur and drastic changes in lighting conditions. Besides, the lack of available ornithopters with suitable payload capacity and the difficulties in the development of these platforms set an additional entry barrier.

This paper presented a perception dataset for bird-scale flapping-wing robots, i.e. ornithopters, recorded in two outdoor and one indoor scenarios. It includes measurements from an event camera, a 
conventional camera, and two IMUs, as well as ground truth data from laser tracker (in outdoor scenarios) and motion capture system (in indoor scenarios).  The provided data include: datasets with agile trajectories that exhibit the ornithopter maneuverability; datasets with smooth trajectories and ArUco markers in the scenario; and datasets for developing and evaluating event-based and visual-based object detection techniques.

The development of perception techniques based on the aforementioned sensors for online execution on board the ornithopter is object of current research. The motion blur and sudden lighting changes observed in the presented datasets recommend the use of event cameras. The vast amount of information captured by the event camera during the ornithopter flight together with the strong payload limitations of these platforms require a significant effort for developing event-based vision methods that can be online and onboard executed. This work intends to set the baseline for future research and contribute to pave the way to develop perception systems that endow the necessary capabilities for ornithopter robots to perceive and interact with the environment.

\section*{Acknowledgment}
The authors would like to thank the support of Rafael Salmoral as multirotor operator, Mart\'in P\'erez at the preparation of the flights, and Miguel P\'erez for data annotation.

\bibliographystyle{IEEEtran}
\bibliography{griffin21dataset}

\end{document}